\pdfoutput=1
\documentclass[11pt]{article}

\usepackage{ACL2023}

\usepackage{times}
\usepackage{latexsym}
\usepackage{amsmath}
\usepackage{graphics}
\usepackage{booktabs}
\usepackage{multirow}
\usepackage{graphicx}
\usepackage{xcolor}
\usepackage{colortbl}
\usepackage{adjustbox}

\usepackage[T1]{fontenc}

\usepackage[utf8]{inputenc}

\usepackage{microtype}

\usepackage{inconsolata}

\title{Bias in Opinion Summarisation from Pre-training to Adaptation: \\ A Case Study in Political Bias}

\author{
\begin{tabular}{c c c}
    Nannan Huang & Haytham Fayek & Xiuzhen Zhang \\
\end{tabular}
\\[5pt]
RMIT University, Australia
\\[5pt]
\texttt{amber.huang@student.rmit.edu.au} \\
\texttt{haytham.fayek@ieee.org} \\
\texttt{xiuzhen.zhang@rmit.edu.au} \\
}
\begin{document}
\maketitle
\begin{abstract}

Opinion summarisation aims to summarise the salient information and opinions presented in documents such as product reviews, discussion forums, and social media texts into short summaries that enable users to effectively understand the opinions therein.
Generating biased summaries has the risk of potentially swaying public opinion.
Previous studies focused on studying bias in opinion summarisation using extractive models, but limited research has paid attention to abstractive summarisation models.
% Limited research has paid attention to studying bias in opinion summarisation using abstractive summarisation models due to the difficulty of measuring bias in this setting.
In this study, using political bias as a case study, we first establish a methodology to quantify bias in abstractive models, then trace it from the pre-trained models to the task of summarising social media opinions using different models and adaptation methods.
We find that most models exhibit intrinsic bias. Using a social media text summarisation dataset and contrasting various adaptation methods, we find that tuning a smaller number of parameters is less biased compared to standard fine-tuning; however, the diversity of topics in training data used for fine-tuning is critical.
% We discuss the implications of our findings for NLP research and propose future directions to combine model performance and fairness measures to assist in designing models with strong performance and minimal bias.

\end{abstract}

\section{Introduction}
Opinion summarisation aims to condense the opinions presented in the source documents into a summary so that readers can effectively comprehend the opinions in the source documents using input data such as product reviews \cite{chu2019meansum, bravzinskas2020unsupervised, hosking-etal-2022-hierarchical}, online discourse using platforms such as Reddit \cite{fabbri-etal-2021-convosumm}, social media text from platforms such as X (formerly known as Twitter) \cite{bilal2022template}, or other types of text containing opinions such as debate \cite{bar2020arguments, bar2020quantitative}.
Applications for this activity vary from tracking customer sentiments to summarising public opinions on political topics.

% Previous studies showed that bias in LLM can propagate to downstream tasks \cite{feng-etal-2023-pretraining}. Research also revealed that using opinionated AI language technologies has the risk of affecting how readers write and think \cite{jakesch2023co}.
% A summarisation model's output will reflect any biases in the data used to train it. If these models produce biased summaries, they face the risk of being used as a tool to persuade and manipulate people's opinions.
% Hence, it is important to understand the biases in the models to ensure they are not used as weapons to sway public opinion or even manipulate the election.

% While LLMs significantly improved performance in different natural-language tasks, numerous studies have shown they have inherent societal bias from the training data \cite{vig2020causal, sheng2019woman, liang2021towards} and propagate to downstream natural-language tasks \cite{feng-etal-2023-pretraining}.
% Extensive amounts of research have studied bias in LLM \cite{vig2020causal, sheng2019woman, liang2021towards} and also showed that bias in LLM can propagate to downstream tasks \cite{feng-etal-2023-pretraining}.
A summarisation model's output will reflect any biases inherited from the training data. Pre-trained language models (PLMs) were exposed to a variety of data that may contain societal bias, which inevitably perpetuates social stereotypes in models \cite{vig2020causal, sheng2019woman, liang2021towards} and can propagate to downstream tasks \cite{feng-etal-2023-pretraining}. Understanding how models inherit societal bias from their training data and how these biases are amplified in downstream tasks is important for designing fair models.
Using opinionated AI language technologies has the risk of affecting how readers read and think \cite{jakesch2023co}. It is important to understand the biases in models to ensure they are not used as weapons to sway public opinion.
% potentially distorting opinions and harming democracy.
% Using biased opinion summarisation models has the risk of producing summaries with distorting opinions rather than the truth, which harms democracy.
% Hence, it is important to understand the biases in the models to ensure they are not used as weapons to sway public opinion.

Prior studies have focused on studying bias in opinion summarisation using extractive models by comparing how contents are extracted 
% from the source documents and their proportion, 
and if they are representing opinions from different social groups in the source documents equally or proportionally \cite{dash2019summarizing, blodgett-etal-2016-demographic, keswani2021dialect, olabisi-etal-2022-analyzing}. 
% whether they are representing opinions from the source documents by having equal or proportional representations of different social groups \cite{dash2019summarizing, blodgett-etal-2016-demographic, keswani2021dialect, olabisi-etal-2022-analyzing}. 
% While the most recent summarisation models are abstractive, making the summaries more readable, coherent, and better able to combine opinions. 
This method is inapplicable to abstractive summarisation models since models generate summaries by rephrasing, making it more challenging to capture and evaluate the opinions represented in the generated documents. 
% Therefore, limited studies have paid attention to examining bias using abstractive summarisation models. 
In addition, fine-tuning abstractive summarisation models is required to build effective summarisation systems.
% abstractive summarisation models require different levels of adaptation, such as fine-tuning, in order to perform well in the task at hand. 
How different adaptation methods introduce bias when summarising social media text has not been studied.

In this study, we use the following definition of fairness: the generated summary must give exposure to the opinions of different social groups equally or proportionally w.r.t. the input documents; more information on this can be found in Section~\ref{fairness_definition}. To address the aforementioned issues, this paper introduces a method using a classifier to identify opinions and a fairness metric to measure bias using abstractive summarisation models to summarise text with opinions, using political bias as the case study.
We further investigate various adaptation methods and the bias introduced, using our method for evaluating bias in abstractive summarisation.
This can be used in conjunction with other performance evaluations to identify models that have good performance while keeping bias to the minimum.
% We develop a classifier to categorise the opinions in the generated sentences, then compare the proportions of the opinions from different social groups against the proportions of the input documents.
% Using this method, we can examine the intrinsic bias in the model when using the pre-trained models directly; this method is also applicable with updated models.
% Our study is two-fold: first, adopting the same notion of fairness as extractive summarisation models, we look into how to measure bias in abstractive summarisation by presenting fair opinions from different political parties. It is unavoidable to use pre-trained models, followed by fine-tuning them to adapt to a downstream task to save and share resources within the community. The second focus of our study is to understand how bias is presented in the pretrained models and how different adaptation methods amplify bias.

% Different tuning methods for abstractive summarisation models are available to adapt to the downstream tasks.
% There is a lack of research on how different methods affect the bias introduced into the model.

We find that different models and their variants express intrinsic bias, and fine-tuning these pre-trained models to summarise social media text amplified the bias. 
In addition, we find that adaptation methods play an important role. We find that tuning a smaller number of parameters using methods such as adapter tuning \cite{houlsby2019parameter} produces less bias compared to standard fine-tuning that updates the entire model. However, the diversity of training data is critical when tuning a model by updating a smaller number of parameters.
Our study is the first of its kind to examine bias using abstractive models to summarise social media text with various adaptation methods.

\section{Related Work}
\subsection{Opinion Summarisation}
Opinion summarisation is a task to summarise user opinions expressed in different online media, such as product reviews, social media conversations, and online discussion forums. There are two primary types of models: extractive --- selecting salient sentences from input documents \cite{mihalcea2004textrank, erkan2004lexrank, inouye2011comparing} and abstractive --- paraphrasing and generating new words and sentences to form the summary \cite{chu2019meansum, bravzinskas2020unsupervised, bravzinskas2021learning}.  
Extensive studies and methods have paid great attention to generating summaries using product reviews where the objective of generating summaries that represent the majority opinions \cite{amplayo-lapata-2020-unsupervised, amplayo-etal-2021-aspect, iso2022comparative, hosking-etal-2023-attributable}. 
% In contrast, the objective of a fair summarisation model is to reflect public opinions equally or proportionally w.r.t the input documents only making models designed to summarise reviews unsuitable to study bias in opinion summarisation.
Abstractive models such as BART \cite{lewis2020bart}, T5 \cite{raffel2020exploring}, and GPT-2 \cite{radford2019language} have led to substantial performance gains in summarisation and also multi-document and opinion summarisation \cite{bravzinskas2022efficient, chen2020multi, johner2021error}. 
However, these models were trained on diverse data sources such as news, books, and discussion forums, where models can inherent societal bias from the training corpus.
Therefore, to use these models directly and adapt them to certain specialised tasks, we need to understand fairness in these models to avoid propagating bias further.
\begin{figure*}[tbh]
    \centering
    \includegraphics[width=10cm]{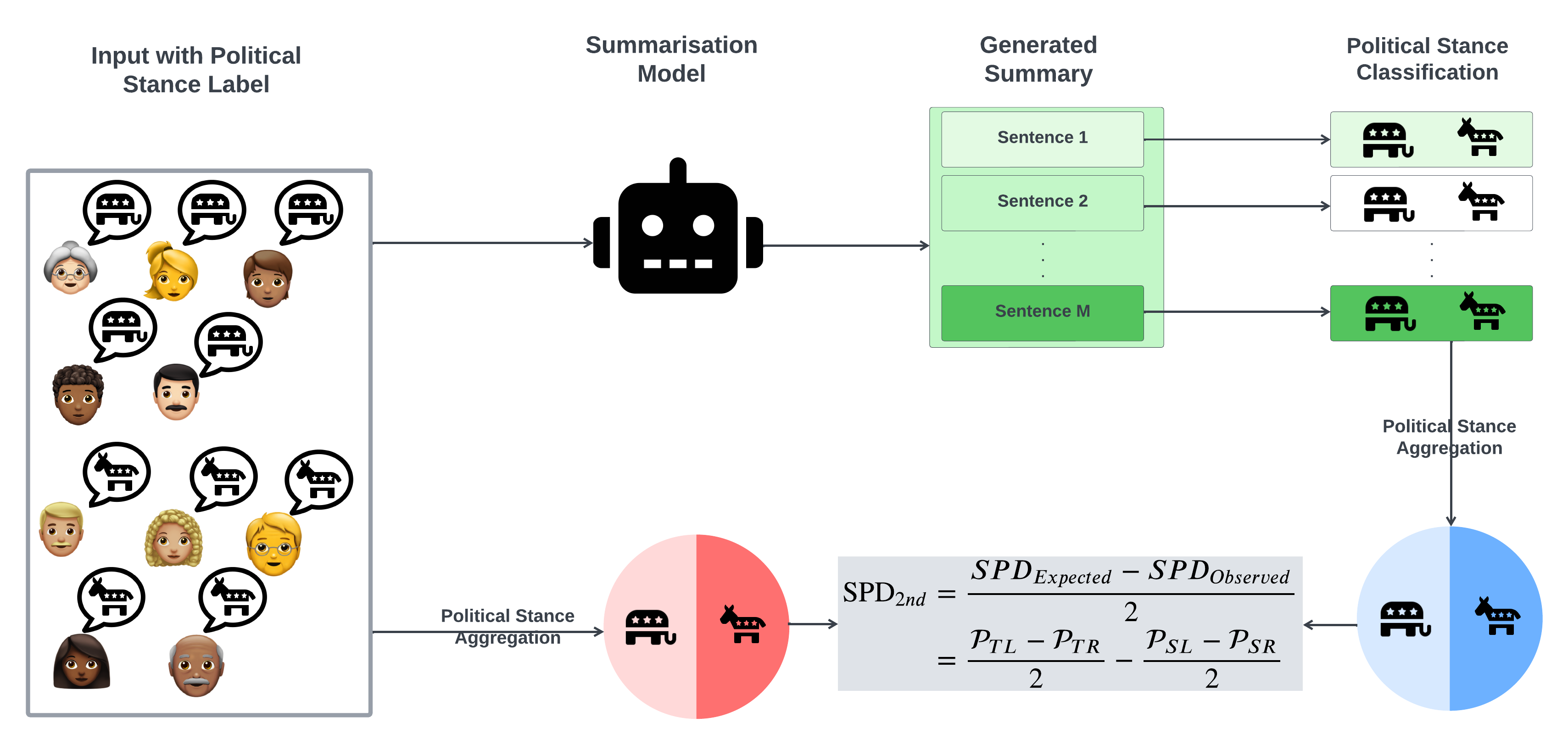}
    \caption{The process of measuring fairness in our study. For the input documents, each tweet has a label indicating the tweet is expressing a left or right-leaning stance. After feeding the input documents to the summarisation models, we split and classify each sentence in the summary to capture its left or right-leaning stance. We aggregate both the source documents and summary sentences on political stances, calculate the Second-order SPD (more detail in Section~\ref{the_metric}), and use it as the fairness measurement.}
    \label{fig:fair}
\end{figure*}

\subsection{Bias in Opinion Summarisation}
Existing studies of bias in opinion summarisation have focused on the perspective of using social attributes of social media users and examining whether the generated summary reflects these groups fairly by selecting text produced by different social groups equally or proportionally using different social attributes such as gender, race and political stance \cite{dash2019summarizing}, dialect \cite{blodgett-etal-2016-demographic, keswani2021dialect, olabisi-etal-2022-analyzing} or opinion diversity \cite{huang-etal-2023-examining}.
% Today, the mainstream summarisation models are abstractive summarisation models \cite{lewis2020bart, raffel2020exploring, radford2019language} using the Transformers architecture \cite{vaswani2017attention}, and generating summaries by rephrasing the original documents. 
% Using such models will generate text that is more fluent, readable, and cohesive compared to the ones generated using extractive summarisation models. 
One limitation of these studies is that they mainly studied bias using extractive summarisation models, whereas the mainstream summarisation models are abstractive summarisation models \cite{lewis2020bart, raffel2020exploring, radford2019language}; in addition, these studies do not focus on the algorithmic bias in the summarisation models. 
In our work, we first focus on studying bias in abstractive summarisation, then we look at how bias is amplified using different adaptation methods through the case study in political bias.

\subsection{Political Bias in Language Models}
Prior work has paid attention to bias in language models. Extensive research has focused on social biases such as gender, race and other social attributes \cite{vig2020causal, sheng2019woman, liang2021towards, ladhak2023pre}. 
% For example, \citet{ladhak2023pre} looked into how bias propagates from pretraining to summarisation using the name and nationality pair. They found that models can hallucinate nationalities based on the provided entity name whereas a fair model should generate output based only on the provided input documents.
It is important to understand political bias in language models because political bias is hard to detect and has a stronger influence on readers than other types of bias \cite{peters2022algorithmic}.
% As a case study, \citet{zhou2023characterizing} has switched the entity name in news using both Trump and Biden while keeping the remaining the same. A fair summarisation model should reflect the sentiment in the input documents. However, they found that the generated summaries about Trump tended to be more negative.
\citet{santurkar2023whose} examined language models' political opinions by comparing their generated output with US survey data and found that language models have opinions on political issues but do not necessarily reflect public opinion.
\citet{feng-etal-2023-pretraining} applied the political compass test to a diverse range of models, then manipulated the political tendency of models by further pretraining them to become left or right-leaning. They found that the bias presented in the pre-trained model propagated to different downstream tasks, and the left-leaning models performed better than the right-leaning models given the same model architecture.
How political bias is presented and propagated has not been studied in the context of opinion summarisation using abstractive summarisation models.

\section{Fairness in Opinion Summarisation}
\label{fairness_definition}
Given a collection of tweets, $\mathcal{T}$, defined as $\mathcal{T} = \{t_0, t_1, t_2,...,t_N\}$. Each tweet $t_i$ has a ground-truth label $y_i \in \mathcal{Y}$ for its political stance, where $\mathcal{Y} = \{y_0, y_1, y_2,...,y_N\}$ represents the label set (left or right-leaning).
Given a set of input tweets, a model would generate a summary $\mathcal{S}$ where each summary consists of a list of sentences defined as $\mathcal{S} = \{s_0,s_1,s_2,...,s_L\}$. Each generated sentence would be classified as left or right-leaning using the trained classification model discussed in Section~\ref{classification}.

Given the set of input tweets $\mathcal{T}$, the proportion of left and right-leaning documents can be represented as $\mathcal{P}_{T L}$ and $\mathcal{P}_{T R}$ respectively. For the generated summary $\mathcal{S}$, the proportion of left and right-leaning sentences can be represented as $\mathcal{P}_{S L}$ and $\mathcal{P}_{S R}$ respectively.
For a model to be considered unbiasedly representing opinions in the provided source documents, it should generate a summary that reflects similar proportions of opinions in the input documents, i.e. $\mathcal{P}_{T L} = \mathcal{P}_{S L}$ and $\mathcal{P}_{T R} = \mathcal{P}_{S R}$, or $\mathcal{P}_{T L}/\mathcal{P}_{T R} = \mathcal{P}_{S L}/\mathcal{P}_{S R}$.

In our study, we focus on evaluating the model's output w.r.t. the input proportions only. We are considering two different input scenarios, namely equal input and skewed input.
The intuition behind and the details of different input proportions in summarising social media text are below:
\begin{itemize}
    \item \textbf{Equal Input}
    In the case of equal input, the input documents contain the same proportion of opinions from different social groups.
    For a model to be considered fair, it should give exposure to opinions from different social groups equally in the generated summary, i.e., if both $\mathcal{P}_{T L}$ and $\mathcal{P}_{T R}$ are 0.5, the generated summary should reflect this by having both $\mathcal{P}_{S L}$ and $\mathcal{P}_{S R}$ equal to 0.5.
    \item \textbf{Skewed Input}
    It is not always practical to have equal distribution in the input documents; instead, they are often proportionally different among different groups. For example, existing studies have shown political parties tweet at different frequencies \cite{pew2020differences, fujiwara2021effect}.
    Given proportional inputs, a fair model should produce summaries that expose opinions from the social groups matching the input documents in proportion, i.e., if $\mathcal{P}_{T L}$ and $\mathcal{P}_{T R}$ are 0.7 and 0.3 respectively, for a model to be considered fair, the generated summary should reflect this by having $\mathcal{P}_{S L}$ and $\mathcal{P}_{S R}$ equal to 0.7 and 0.3 respectively, in this case, a model having $\mathcal{P}_{S L}$ and $\mathcal{P}_{S R}$ both equal to 0.5 would not be considered fair.
\end{itemize}
% Following the notion of fairness by \citet{dash2019summarizing}, we are measuring two types of fairness in our paper: equal representation and proportional representation. In our study, we focus on evaluating the model's output w.r.t. the input proportions only. The detail of our definition of fair summarisation using social media text is below:
% \begin{itemize}
%     \item \textbf{Equal Representation}
%     The notion of equal representation requires the summarisation model to expose opinions from different social groups equally in the generated summary when the input documents are equally distributed, i.e., if both $\mathcal{P}_{T L}$ and $\mathcal{P}_{T R}$ are 0.5, for a model to be considered fair, the generated summary should reflect this by having both $\mathcal{P}_{S L}$ and $\mathcal{P}_{S R}$ equal to 0.5.
%     \item \textbf{Proportional Representation}
%     It is not always practical to have equal distribution in the input documents; instead, they are often proportionally different among different groups. The summary produced by the model should match the input documents in proportion in order to reflect this, i.e., if $\mathcal{P}_{T L}$ and $\mathcal{P}_{T R}$ are 0.7 and 0.3 respectively, for a model to be considered fair, the generated summary should reflect this by having $\mathcal{P}_{S L}$ and $\mathcal{P}_{S R}$ equal to 0.7 and 0.3 respectively, in this case, a model having $\mathcal{P}_{S L}$ and $\mathcal{P}_{S R}$ both equal to 0.5 would not be considered fair.
% \end{itemize}

We evaluate fairness in models based on the idea that the generated summary should give exposure to opinions representing different social groups w.r.t. the input only. More details on the metric we are adapting using these notions for evaluation can be found in Section~\ref{the_metric}. Note that our notion of fairness can be broadly applicable to the summarisation of different types of opinions in other genres, such as positive or negative opinions on specific issues.

\section{Methodology}
We formulate our problems in three steps. We first use a classification model to determine whether the sentences in the generated summary represent opinions from left or right-leaning groups. Then, using the metric to assess whether a model contains left or right-leaning bias and quantifying the severity by comparing the generated summaries w.r.t. the input documents. 
% These lay the groundwork for us to examine bias using abstractive summarisation models.
The overall process of measuring bias is visualised in Figure~\ref{fig:fair}.
As demonstrated by earlier research \cite{han2021robust, li2021prefix, kirichenko2022last, chen2023unisumm}, tuning a smaller set of parameters can result in more robust performance than standard fine-tuning. 
% Several studies revealed that tuning a smaller set of parameters has the benefit of building more robust models \cite{han2021robust, li2021prefix, kirichenko2022last, chen2023unisumm}.
However, there is a lack of research on how different methods affect the bias introduced into the model.
Finally, we examine how different approaches amplify bias as compared to standard fine-tuning. 
% The aim is to determine the most suitable adaptation method that retains good model performance while minimising bias introduced to the model. 
% A more detailed discussion of how the classification model is trained, how we examine bias, the abstractive summarisation models, and the adaptation methods we use in our study can be found below.

% Lastly, we aim to determine the best strategy that retains good model performance while minimising bias by applying several models and fine-tuning strategies. A more detailed discussion of how we examine bias in abstractive summarisation, models and fine-tuning strategies can be found below.

\subsection{Classification of Political Stance}
\label{classification}
% Existing studies evaluate bias in extractive models by comparing the selected documents' labels and calculating their proportions in the generated summary \cite{dash2019summarizing, keswani2021dialect, olabisi-etal-2022-analyzing}. This method is inapplicable when using abstractive models since the generated sentences are rephrased from the original documents rather than directly selected from the input documents. This makes it impossible to evaluate fairness using abstractive models using the input labels directly. 

% To overcome this challenge, we first train a model that can classify the generated sentences according to the political stances. 
We use a RoBERTa \cite{liu2019roberta} further pretrained on the tweet dataset \cite{barbieri2020tweeteval}\footnote{\url{https://huggingface.co/cardiffnlp/twitter-roberta-base}.}
and then fine-tuned using the political partition of the dataset provided by \citet{dash2019summarizing}. We randomly divided the dataset into 80\% and 20\% for training and testing respectively. For model development, we use 70\% of the training subset for training and 30\% for validation.
Note that since our primary focus is on text with opinions, therefore, we are only using the left and right-leaning tweets from the dataset. Thus, our tweet political stance classification model is a further pre-trained RoBERTa fine-tuned with standard cross entropy loss to do binary classification of the political stance label (left or right). 
Each tweet $t_i$ is associated with a ground-truth label $y_i \in \mathcal{Y}$, where $\mathcal{Y}$ represents the label set (left or right; 2 classes).
\begin{align}
    v_i &= \text{RoBERTa}([\text{CLS}] \oplus t_i )\label{eqn:roberta} \\ 
    \hat{y_i} &= \text{softmax} (W v_i + b)  \label{eqn:softmax}
\end{align}
Detail of the training process can be found in Section~\ref{sec:experiment_details}. The average accuracy and macro F1 scores of the model are 0.9162 and 0.9031 respectively. The majority of the input documents contained only a single sentence. We, therefore, treat each sentence in the generated summaries as a tweet and apply the classifier to retrieve opinions in the generated summary.

Model generated summaries often consist of compound sentences that contain opposing opinions due to their abstract nature. To overcome this issue, we first use ChatGPT (we use OpenAI's ChatGPT API (gpt-3.5-turbo-0301) for our experiments) to split these compound summaries into sentences containing only a single opinion by prompting "Split the following sentences into simple propositions without introducing new information, do it sentence by sentence: \textbackslash n\textbackslash n  Sentences:". Then apply the classifier to each of these sentences. 
Note that the summarisation dataset provided by \citet{bilal2022template} uses a template to represent opinions with varying proportions; in our evaluation for sentences containing "the minority", we assign a weight that is half that of the other sentences.
By taking quantitative factors into account, these weights are used to determine the proportion of political stances in the generated summaries.

\subsection{Measuring Bias in Abstractive Summarisation}
\label{the_metric}
Calculating the proportion of left and right-leaning in the input tweets and summary provides a set of opinion distributions in both the source documents and the summary.
% , denoted as $D(T)$ and $D(S)$ respectively. 
To answer the question of whether the generated summary exposes opinions in the input documents equally or proportionally, a similarity measure over pairs of such distributions is required.

Even though we can compare two distributions using any distributional divergence, there are some intricacies in the differences between the two distributions that we would like to capture. In particular, which side is a biased model more likely to give exposure to? This means that divergence measures such as the Kullback-Liebler or 1-Wasserstein distance are insufficient as they are deemed not directional.

We thus turn to a fairness notion called statistical parity that is used to evaluate fairness in machine learning models and decision-making procedures \cite{barocas-hardt-narayanan}. A measure based on statistical parity called the Statistical Parity Difference (SPD) measures the difference in the proportion of favourable outcomes between different groups in a model's predictions; in our case, a model includes more opinions representing one group over the other.
Typically, this measure must be equal to zero to be fair. However, since we are also capturing situations that are not equally distributed, we hence build on the measurement to meet our requirement, namely \textit{Second-order SPD}. We have the \textit{Expected SPD} that is calculated using the input distribution and the \textit{Observed SPD} that is calculated using the generated summary distribution. 
The Second-order SPD is the difference between the Expected and the Observed SPD, which reflects both the magnitude and direction of the bias. We standardise the formula so that it ranges from -1 to 1, where the absolute value of the metric indicates how severe the bias is, and the sign represents which side the model leans more towards. For example, in the case of political bias, a value of -1 means absolute bias towards the left and 1 means absolute bias towards the right. We examine and discuss the necessity of adopting Second-order SPD rather than SPD in Appendix~\ref{sec:spd_and_second_order_spd}. The formula of Second-order SPD can be found below:
\begin{align} \label{spd_difference_with_denominator}
\mathrm{SPD}_{2nd} &= \frac{{SPD}_{Expected} - {SPD}_{Observed}}{2} \nonumber \\
&= \frac{\mathcal{P}_{T L} - \mathcal{P}_{T R}}{2} - \frac{\mathcal{P}_{S L} - \mathcal{P}_{S R}}{2}
\end{align}

\begin{align} \label{spd_difference_range}
\mathrm{SPD}_{2nd} \in [-1, 1]
\end{align}

In our experiments, we report the average Second-order SPD as the overall fairness measurement for each model with different input proportions.
% \begin{equation} \label{spd_difference}
% \begin{split} 
% \mathrm{SPD}_{Difference} & = {SPD}_{Expected} \\
% & - {SPD}_{Observed} \\
% & = (\mathcal{P}_{T L} - \mathcal{P}_{T R}) \\
% & - (\mathcal{P}_{S L} - \mathcal{P}_{S R})
% \end{split}
% \end{equation}

\subsection{Models}
\label{models}
We use existing state-of-the-art abstractive summarisation models with different architectures and variants in our study. Including encoder-decoder models BART \cite{lewis2020bart} and T5 \cite{raffel2020exploring} and also decoder only model GPT-2 \cite{radford2019language}. A more detailed discussion of each model can be found below. We use these models because they have similar model sizes across the variants; in addition, they are open-source, which allows us to investigate algorithmic bias using the different adaptation methods we mention in Section~\ref{adaptation_methods}.
All models are implemented in PyTorch using the HuggingFace library.\footnote{\url{https://github.com/huggingface}}

\begin{itemize}
    \item \textbf{BART} \cite{lewis2020bart} is an encoder-decoder model with a bidirectional encoder and a left-to-right decoder.
    Pretrained using document rotation, sentence permutation and a novel in-filling technique by replacing a span of text with a single mask token. 
    We use the BART base and BART large in our experiments.
    \item \textbf{T5} \cite{raffel2020exploring} is a text-to-text model with an encoder-decoder architecture that has been pretrained on a multi-task environment utilising both supervised and unsupervised training, where the tasks are transformed into a set of input-output text pairs. We use T5 small, T5 base and T5 large in our experiments. 
    \item\textbf{GPT-2} \cite{radford2019language} is a decoder only model that was self-supervisedly pretrained on a large corpus of English data. The model was pretrained on conditional generation, and it is known for producing texts in response to a prompt. We use Distilled-GPT2, GPT-2, GPT-2 Medium and GPT-2 Large in our experiments.

\end{itemize}

\subsection{Adaptation Methods}
\label{adaptation_methods}
% In order to create the summary, extractive summarisation models use word or sentence embedding to calculate the similarity between sentences and weight each sentence's importance by its similarity with others; the ones with the highest weights are selected. In most cases, extractive summarisation models can adapt to summarising various types of text without needing to be updated. 
Different from extractive summarisation models, abstractive summarisation models generate the summary to cover key information in the input documents by rephrasing. To achieve this, a certain level of model tuning is required, and different adaptation methods can be applied.
We are using the following adaptation methods on all models mentioned in Section~\ref{models}: 
% more detail about each adaptation method can be found below:

\begin{itemize}
    \item \textbf{Standard fine-tune}
    the models mentioned in Section~\ref{models} are further trained on a dataset to adapt to the specific task, during this step, the model's parameters are all updated to better adapt to the task at hand.
    
    \item \textbf{Adapter tuning}
    instead of updating all parameters in a model, adapter tuning introduces adapter layers in the original model and only updates parameters in these layers \cite{houlsby2019parameter}. This method was introduced for more efficient learning and also to mitigate potential catastrophic forgetting issues. The adapter-based models we use in this work are from AdapterHub \cite{pfeiffer2020adapterhub}.\footnote{\url{https://docs.adapterhub.ml/}}
    
    \item\textbf{Prefix-tuning} \cite{li2021prefix} is an additive method where the beginning of the input (prefix), is connected to a series of continuous vectors that are specific to the task at hand. In every layer of the model, the hidden states are appended with the prefix parameters; upon tuning, only the prefix parameters will be updated. The tokens of the input sequence can still attend to the prefix as virtual tokens. Our implementation of prefix-tuning is using the PEFT library from HuggingFace.\footnote{\url{ https://huggingface.co/docs/peft/index }} For models mentioned in Section~\ref{models} we introduce 200 virtual tokens so that models have a similar percentage of parameter updates compared to the adapter tuning.

    \item\textbf{Last decoder layer tuning} we freeze all pretrained parameters for the models stated in Section~\ref{models} with the exception of the final decoder layer. This would only update the final layer of the decoder, leaving the other layers of the model unchanged.
\end{itemize}

\nocite{Ando2005,augenstein-etal-2016-stance,andrew2007scalable,rasooli-tetrault-2015,goodman-etal-2016-noise,harper-2014-learning}

\section{Results and Discussion}
\subsection{Datasets}
In this study, we use the tweet summarisation dataset provided by \citet{bilal2022template} to fine-tune models for social media text summarisation.
They published 2,214 tweet clusters that had been manually identified as coherent, covering COVID-19 (2021-2022) and political (2014-2016) topics. 
For each cluster, there are on average 30 tweets posted by different users on the same day and discuss the same subtopic. 
% We randomly select 80\% of the clusters for training and 20\% for validation then test the model's performance using the provided test set. 
% Models mentioned in Section~\ref{models} are tuned using this dataset with the adaptation mentioned in Section~\ref{adaptation_methods}.
The input documents are first sorted by time and then truncated to fit the maximum input length of 1024 for all models. Following \citet{bilal2022template} we limit the abstract summarisation models word limit to the generated summary within [90\%, 110\%] of the gold standard length.
We trained the models using the provided training set: 80\% for training and 20\% for evaluation, with a batch size of 16, for 10 epochs with early stopping, with a learning rate resulting in the lowest validation loss. Then evaluated on the provided test set.

To test whether a model has political bias when summarising social media text, we use the political partition of the dataset provided by \citet{dash2019summarizing}. The dataset contains 2,120 tweets, out of which 1,309 (61.74\%) are right-leaning, 658 (31.04\%) tweets are left-leaning, and the remaining 153 (7.22\%) are neutral tweets. 
As mentioned earlier, we focus on text with opinions only; therefore, when generating summaries, we exclude the neutral tweets similarly to the classification task mentioned in Section~\ref{classification}.

% Note that, since we are exclusively focused on political stances and creating opinionated text as the output, we are not using the neutral tweets in this instance for both summarisation and the classification task mentioned in Section~\ref{classification}.

Recall that we use different input proportions to examine model fairness, we generate the testing dataset as follows:
for equal input, we select 50\% of tweets from both political stances. We have two scenarios for skewed input: one with more left-leaning tweets (where 75\% of the inputs are left-leaning and 25\% are right-leaning) and one with more right-leaning tweets (where 25\% of the inputs are left-leaning and 75\% are right-leaning).
For each scenario, we create 100 test inputs with 20 tweets each to ensure it is within maximum input length limit. The purpose of this is to determine if the model can fairly represent both sides given an equal input; in the case of skewed inputs, whether the model can reflect the stances proportionally.
A fair model should generate summaries exposing opinions from different social groups w.r.t. the opinion proportions presented in the source documents only.

% addressing reviewer 2's 4th and reviewer 3's 1b comment
In summary, we first adapt models to summarise social media, test model performance using data provided by \citet{bilal2022template}, and report model performance using ROUGE scores. Subsequently, we utilise the dataset from \citet{dash2019summarizing} to assess political bias. The process is two-fold: we first train a classification model using the political stance labels provided by \citet{dash2019summarizing}; the classification model is then used to classify the generated output from the summarisation models at sentence level. Next, we apply the tuned summarisation models to the handcrafted input data provided by \citet{dash2019summarizing} by adjusting the input proportion. This process aimed to rigorously assess the model's ability in representing different input proportions.
\subsection{Intrinsic Bias of Different Models}
 \begin{table}[tbh!]
\resizebox{\columnwidth}{!}{%
\begin{tabular}{l c c c}
\toprule
Model & $\mathrm{SPD}_{2nd}$-Equal & $\mathrm{SPD}_{2nd}$-Left & $\mathrm{SPD}_{2nd}$-Right \\ \midrule
BART Base & -0.0262 & 0.1219 & -0.2285 \\ 
BART Large & -0.0240 & 0.0708 & -0.2279 \\ 
Distil GPT-2 & -0.1154 & 0.0321 & -0.3520 \\ 
GPT-2 & -0.0345 & -0.0115 & -0.2839 \\ 
GPT-2 Medium & -0.0162 & -0.0160 & -0.2619 \\ 
GPT-2 Large & 0.0012 & -0.0345 & -0.2913 \\ 
T5 Small & -0.0415 & 0.0424 & -0.1957 \\
T5 Base & -0.1385 & -0.0390 & -0.2479 \\
T5 Large & -0.0160 & 0.1205 & -0.2698 \\ \bottomrule
\end{tabular}%
}
  \caption{Intrinsic bias in different models under zero-shot setting for summary generation. The Second-order SPD ($\mathrm{SPD}_{2nd}$) is reported for measuring the fairness of models using different input proportions (equal, more left-leaning, and more right-leaning). Model performance can be found in Table~\ref{zero_shot_performance} in Appendix~\ref{sec:zero_short_model_performance}.
  }
  \label{zero_shot_table}
\end{table}

\begin{table*}[tbh!]
\centering
\begin{adjustbox}{max width=0.7\textwidth}
% \resizebox{\textwidth}{!}{%
\begin{tabular}{|c|c|c|c|c|c|c|c|}
\hline
\textbf{Model} & \textbf{Adaptation Methods} & \multicolumn{1}{l|}{\textbf{ROUGE-1}} & \multicolumn{1}{l|}{\textbf{ROUGE-2}} & \multicolumn{1}{l|}{\textbf{ROUGE-L}} & \multicolumn{1}{l|}{\textbf{$\boldsymbol{\mathrm{SPD}_{2nd}}$-Equal}} & \multicolumn{1}{l|}{\textbf{$\boldsymbol{\mathrm{SPD}_{2nd}}$-Left}} & \multicolumn{1}{l|}{\textbf{$\boldsymbol{\mathrm{SPD}_{2nd}}$-Right}} \\ \hline
\multirow{4}{*}{\textbf{BART Base}} & \textbf{Standard} & 32.02 & 12.02 & 22.73 & -0.2582 (4) & -0.1111 (3) & -0.4617 (4) \\
 & \textbf{Adapter} & 31.88* & 12.21* & 22.80* & -0.0530 (3) & -0.0090 (2) & -0.1106 (2) \\ 
 & \textbf{Prefix} & 29.37 & 9.89 & 20.00 & 0.0502 (2) & 0.1666 (4) & \textbf{-0.1083 (1)} \\ 
 & \textbf{Last Layer} & 29.82 & 10.39 & 20.56 & \textbf{-0.0470 (1)} & \textbf{0.0247 (1)} & -0.2370 (3) \\ \hline
\multirow{4}{*}{\textbf{BART Large}} & \textbf{Standard} & 31.20 & 11.63 & 22.06 & -0.2895 (4) & -0.1582 (3) & -0.4664 (4) \\ 
 & \textbf{Adapter} & 31.95* & 12.22* & 22.73* & \textbf{-0.0520 (1)} & \textbf{0.0518 (1)} & \textbf{-0.1869 (1)} \\
 & \textbf{Prefix} & 26.87 & 9.01 & 16.80 & -0.0835 (2) & 0.1735 (4) & -0.2004 (2) \\ 
 & \textbf{Last Layer} & 29.98 & 10.00 & 20.33 & -0.1648 (3) & -0.0816 (2) & -0.3906 (3) \\ \hline
\multirow{4}{*}{\textbf{Distil GPT-2}} & \textbf{Standard} & 21.76 & 5.78 & 16.44 & -0.2788 (3) & -0.0829 (3) & -0.4766 (3) \\
 & \textbf{Adapter} & 21.12* & 4.95* & 14.95* & \textbf{-0.1568 (1)} & \textbf{0.0347 (1)} & \textbf{-0.3307 (1)} \\
 & \textbf{Prefix} & 10.39 & 3.02 & 8.21 & -0.3532 (4) & -0.1368 (4) & -0.5357 (4) \\
 & \textbf{Last Layer} & 12.83 & 2.86 & 9.63 & -0.2110 (2) & -0.0673 (2) & -0.3812 (2) \\ \hline
\multirow{4}{*}{\textbf{GPT-2}} & \textbf{Standard} & 22.74 & 5.93 & 16.05 & -0.2264 (4) & -0.0883 (3) & -0.4768 (4) \\
 & \textbf{Adapter} & 21.34* & 4.97* & 14.84* & -0.1331 (2) & \textbf{0.0272 (1)} & -0.3889 (3) \\
 & \textbf{Prefix} & 10.13 & 2.61 & 7.99 & \textbf{-0.0833 (1)} & 0.1136 (4) & \textbf{-0.3611 (1)} \\
 & \textbf{Last Layer} & 19.23 & 4.22 & 13.87 & -0.1549 (3) & -0.0569 (2) & -0.3634 (2) \\ \hline
\multirow{4}{*}{\textbf{GPT-2 Medium}} & \textbf{Standard} & 23.39 & 6.43 & 16.94 & -0.2262 (4) & \textbf{0.0077 (1)} & -0.4227 (3) \\
 & \textbf{Adapter} & 22.46* & 6.12* & 16.20* & \textbf{-0.1421 (1)} & 0.0291 (3) & -0.3844 (2) \\
 & \textbf{Prefix} & 16.78 & 5.78 & 12.80 & -0.1525 (2) & 0.0638 (4) & \textbf{-0.3711 (1)} \\
 & \textbf{Last Layer} & 19.37 & 3.61 & 13.50 & -0.1835 (3) & -0.0165 (2) & -0.4478 (4) \\ \hline
\multirow{4}{*}{\textbf{GPT-2 Large}} & \textbf{Standard} & 24.58 & 8.13 & 18.45 & -0.2030 (4) & -0.0225 (3) & -0.3490 (4) \\
 & \textbf{Adapter} & 23.52* & 6.62* & 16.30* & -0.1715 (3) & -0.0172 (2) & \textbf{-0.2951 (1)} \\
 & \textbf{Prefix} & 12.54 & 4.25 & 9.42 & \textbf{-0.0670 (1)} & \textbf{0.0166 (1)} & -0.3038 (2) \\
 & \textbf{Last Layer} & 19.26 & 5.18 & 14.24 & -0.1403 (2) & -0.0554 (4) & -0.3425 (3) \\ \hline
\multirow{4}{*}{\textbf{T5 Small}} & \textbf{Standard} & 27.75 & 9.74 & 19.52 & -0.1891 (2) & -0.0672 (2) & \textbf{-0.3129 (1)} \\
 & \textbf{Adapter} & 24.89 & 9.05 & 17.42 & -0.3464 (4) & -0.1681 (4) & -0.5191 (4) \\ 
 & \textbf{Prefix} & 28.10* & 9.56* & 19.03 & -0.2494 (3) & -0.0983 (3) & -0.4784 (3) \\
 & \textbf{Last Layer} & 27.86 & 9.31 & 19.28* & \textbf{-0.1831 (1)} & \textbf{-0.0485 (1)} & -0.3791 (2) \\ \hline
\multirow{4}{*}{\textbf{T5 Base}} & \textbf{Standard} & 29.86 & 9.82 & 20.49 & -0.1297 (3) & \textbf{0.0338 (1)} & -0.2512 (2) \\
 & \textbf{Adapter} & 27.94* & 10.17* & 20.19* & \textbf{0.0284 (1)} & 0.1397 (4) & \textbf{-0.1263 (1)} \\
 & \textbf{Prefix} & 25.40 & 9.03 & 18.11 & -0.2150 (4) & -0.0593 (3) & -0.3530 (4) \\ 
 & \textbf{Last Layer} & 26.49 & 7.85 & 17.38 & -0.1293 (2) & 0.0430 (2) & -0.2913 (3) \\ \hline
\multirow{4}{*}{\textbf{T5 Large}} & \textbf{Standard} & 31.08 & 11.52 & 22.20 & -0.1211 (3) & \textbf{0.0072 (1)} & -0.2951 (2) \\
 & \textbf{Adapter} & 30.34* & 11.30* & 21.85* & -0.1207 (2) & 0.0133 (2) & \textbf{-0.2069 (1)} \\
 & \textbf{Prefix} & 26.44 & 9.66 & 18.91 & -0.4917 (4) & -0.2427 (4) & -0.7376 (4) \\
 & \textbf{Last Layer} & 22.80 & 7.58 & 16.25 & \textbf{-0.0808 (1)} & 0.0570 (3) & -0.3522 (3) \\ \hline
\end{tabular}%
% }
\end{adjustbox}
  \caption{Results of model performance and fairness evaluation. We highlight the adaptation methods apart from standard fine-tuning with the highest ROUGE score using *. We report Second-order SPD ($\mathrm{SPD}_{2nd}$)  with different input proportions (equal, more left-leaning, and more right-leaning), the lowest absolute values are bolded and the ranking compared between adaptation methods is provided inside the brackets. 
  }
  \label{result_table}
\end{table*}
% Different from extractive models, abstractive models need to be fine-tuned to be effectively used for summarisation. 
% Pre-trained LLMs were exposed to a variety of data that may contain societal bias, which inevitably perpetuates social stereotypes in models \cite{vig2020causal, sheng2019woman, liang2021towards}. Understanding how models inherit societal bias from their training data and how these biases are amplified in their downstream tasks is important for designing a fair model.
Since we would not anticipate a model to prefer one side over the other by exposing more opinions representing a particular social group, we are using the term intrinsic bias to denote political bias in social media text summarisation in pre-trained models.
We measure the intrinsic bias by looking at the bias expressed when applying models in a zero-shot setting.

% The result of the model performance can be found in Table~\ref{}. Compared to fine-tuned models, zero-shot performance is not as good. However, this is expected since models are trained on different types of data and a certain level of fine-tuning is expected to have good performance for a particular dataset.

The result of intrinsic bias can be found in Table~\ref{zero_shot_table}. A fair model should have a close to zero absolute value of Second-order SPD; negative values indicate including more left-leaning information than it should, and positive values indicate including more right-leaning information than the model should. A model should achieve a close to zero reading for all three input proportions to indicate complete fairness by reflecting political stances w.r.t. the input only. The Second-order SPD ($\mathrm{SPD}_{2nd}$) is reported for measuring the fairness of models using different input proportions (equal, more left-leaning, and more right-leaning), and calculated by averaging across test instances.
We find that most models can fairly represent the input political stances when the provided inputs are balanced or contain more left-leaning information. 
However, when providing more right-leaning input, all models failed to expose opinions proportionally in the generated summaries.
Overall, models are better at exposing left-leaning opinions than right-leaning opinions, indicating models are expressing left-leaning bias, which is consistent with the zero-shot findings of \citet{feng-etal-2023-pretraining}.
Through examining models of different sizes, we have not found a clear relationship between model size and the political bias expressed by models.

% From the result, we can see that other than Distil GPT-2 and T5 base, most models can fairly represent opinions from both parties when the input is balanced. When including more left-leaning tweets, most models can reflect the input being more left-leaning by having an SPD difference close to zero, but BART base and T5 large tend to include more right-leaning information. When presenting more right-leaning information, all models failed to represent this situation but included more left-leaning information than they were provided. We have not found a clear relationship between model size and the political bias expressed by models when comparing models with the same architecture but with different model sizes.

% Overall, from the standpoint of intrinsic bias, most models are left-leaning. With equally distributed input, most models can capture both sides equally. When the inputs are proportionate, the models fairly represented left-leaning information when there were more left-leaning inputs but failed to express more right-leaning information when the inputs contained more right-leaning information.
% From our observation, the size of the model has no obvious effect on a model's intrinsic bias.

\subsection{Different Adaptation Methods and Bias}
\label{sec:diff_adap_methods_bias}
\begin{table*}[tbh!]
\centering
\begin{adjustbox}{max width=0.7\textwidth}
% \resizebox{\textwidth}{!}{%
\begin{tabular}{|c|c|c|c|c|c|c|c|}
\hline
\multicolumn{2}{|c|}{  } & \multicolumn{3}{|c|}{ \textbf{COVID-19} } & \multicolumn{3}{|c|}{ \textbf{Elections} } \\
\hline
\textbf{Model} & \textbf{Adaptation Methods} & \multicolumn{1}{l|}{\textbf{$\boldsymbol{\mathrm{SPD}_{2nd}}$-Equal}} & \multicolumn{1}{l|}{\textbf{$\boldsymbol{\mathrm{SPD}_{2nd}}$-Left}} & \multicolumn{1}{l|}{\textbf{$\boldsymbol{\mathrm{SPD}_{2nd}}$-Right}} & \multicolumn{1}{l|}{\textbf{$\boldsymbol{\mathrm{SPD}_{2nd}}$-Equal}} & \multicolumn{1}{l|}{\textbf{$\boldsymbol{\mathrm{SPD}_{2nd}}$-Left}} & \multicolumn{1}{l|}{\textbf{$\boldsymbol{\mathrm{SPD}_{2nd}}$-Right}} \\ \hline
\multirow{4}{*}{\textbf{BART Base}} & \textbf{Standard} & -0.1155 (2) & -0.0428 (2)  & -0.2316 (3)  & -0.1429 (2)  & \textbf{0.0020 (1)}  & -0.2971 (2) \\
 & \textbf{Adapter} & -0.2063 (3) & -0.0852 (4)  & -0.2195 (2)  & \textbf{-0.0819 (1)}  & -0.0258 (4)  & \textbf{-0.1875 (1)} \\
 & \textbf{Prefix} & -0.2360 (4) & \textbf{-0.0260 (1)}  & -0.3836 (4)  & -0.2337 (4)  & -0.0244 (3)  & -0.3845 (4) \\ 
 & \textbf{Last Layer} & \textbf{0.0042 (1)} & 0.0844 (3)  & \textbf{-0.2167 (1)}  & -0.1489 (3)  & 0.0234 (2)  & -0.3697 (3) \\ \hline
\multirow{4}{*}{\textbf{BART Large}} & \textbf{Standard} & -0.0714 (2) & \textbf{0.0179 (1)}  & -0.2663 (3)  & -0.0751 (2)  & \textbf{0.0054 (1)}  & -0.2661 (3) \\ 
 & \textbf{Adapter} & -0.1173 (4) & 0.0252 (3)  & -0.2935 (4)  & -0.1634 (3)  & -0.0557 (4)  & -0.2537 (2) \\
 & \textbf{Prefix} & -0.0824 (3) & 0.0874 (4)  & -0.2377 (2)  & -0.2236 (4)  & -0.0556 (3)  & -0.4108 (4) \\ 
 & \textbf{Last Layer} & \textbf{-0.0408 (1)} & -0.0248 (2)  & \textbf{-0.1965 (1)}  & \textbf{-0.0302 (1)}  & 0.0093 (2)  & \textbf{-0.2036 (1)} \\ \hline
\multirow{4}{*}{\textbf{Distil GPT-2}} & \textbf{Standard} & -0.2687 (4) & -0.0869 (4)  & -0.4889 (3)  & -0.2125 (4)  & -0.0326 (3)  & -0.3560 (3) \\
 & \textbf{Adapter} & \textbf{-0.1270 (1)} & \textbf{0.0043 (1)}  & \textbf{-0.3526 (1)}  & -0.0623 (2)  & 0.0235 (2)  & \textbf{-0.2639 (1)} \\
 & \textbf{Prefix} & -0.2532 (3) & -0.0591 (2)  & -0.5371 (4)  & \textbf{-0.0301 (1)}  & 0.1777 (4)  & -0.2655 (2) \\
 & \textbf{Last Layer} & -0.2478 (2) & -0.0659 (3)  & -0.4677 (2)  & -0.1784 (3)  & \textbf{-0.0151 (1)}  & -0.4136 (4) \\ \hline
\multirow{4}{*}{\textbf{GPT-2}}  & \textbf{Standard} & -0.2539 (3) & -0.0874 (4)  & -0.4398 (3)  & -0.2112 (4)  & -0.0233 (2)  & -0.3379 (2) \\
 & \textbf{Adapter} & \textbf{-0.1390 (1)} & -0.0177 (2)  & \textbf{-0.3623 (1)}  & -0.1870 (3)  & \textbf{-0.0197 (1)}  & -0.3838 (4) \\
 & \textbf{Prefix} & -0.3159 (4) & \textbf{-0.0020 (1)}  & -0.5128 (4)  & -0.1586 (2)  & 0.0278 (3)  & -0.3437 (3) \\
 & \textbf{Last Layer} & \textbf{-0.1390 (1)} & -0.0212 (3)  & -0.4200 (2)  & \textbf{-0.0942 (1)}  & -0.0296 (4)  & \textbf{-0.3212 (1)} \\ \hline
\multirow{4}{*}{\textbf{GPT-2 Medium}} & \textbf{Standard} & -0.2626 (3) & -0.0663 (4)  & -0.4830 (4)  & -0.0674 (2)  & 0.0617 (4)  & \textbf{-0.2398 (1)} \\
 & \textbf{Adapter} & \textbf{-0.1424 (1)} & \textbf{-0.0092 (1)} & \textbf{-0.3500 (1)}  & -0.1857 (4)  & -0.0067 (2)  & -0.3759 (4) \\
 & \textbf{Prefix} & -0.3048 (4) & 0.0188 (3)  & -0.4714 (3)  & -0.1797 (3)  & 0.0505 (3)  & -0.3100 (2) \\
 & \textbf{Last Layer} & -0.2492 (2) & -0.0169 (2)  & -0.4713 (2)  & \textbf{-0.0403 (1)}  & \textbf{0.0059 (1)}  & -0.3489 (3) \\ \hline
\multirow{4}{*}{\textbf{GPT-2 Large}} & \textbf{Standard} & -0.1733 (4) & -0.0420 (3)  & -0.4285 (4)  & -0.1497 (4)  & \textbf{0.0157 (1)}  & -0.2229 (2) \\
 & \textbf{Adapter} & -0.1101 (2) & 0.0362 (2)  & -0.3190 (2)  & -0.1191 (2)  & 0.0496 (3)  & -0.2734 (3) \\
 & \textbf{Prefix} & \textbf{0.0158 (1)} & 0.2137 (4)  & \textbf{-0.2639 (1)}  & -0.1245 (3)  & 0.0164 (2)  & -0.4310 (4) \\
 & \textbf{Last Layer} & -0.1212 (3) & \textbf{-0.0237 (1)}  & -0.3683 (3)  & \textbf{-0.0023 (1)}  & 0.0684 (4)  & \textbf{-0.1272 (1)} \\ \hline
 \multirow{4}{*}{\textbf{T5 Small}} & \textbf{Standard} & -0.1438 (2) & -0.0277 (2)  & \textbf{-0.3182 (1)}  & \textbf{-0.1076 (1)}  & \textbf{0.0075 (1)}  & -0.2919 (2)\\
 & \textbf{Adapter} & -0.3145 (4) & -0.1613 (4)  & -0.5335 (4)  & -0.1918 (4)  & -0.0365 (3)  & -0.3535 (3)\\
 & \textbf{Prefix} & -0.2264 (3) & -0.0817 (3)  & -0.4268 (3)  & -0.1701 (2)  & -0.0086 (2)  & \textbf{-0.2828 (1)}\\
 & \textbf{Last Layer} & \textbf{-0.1168 (1)} & \textbf{-0.0008 (1)}  & -0.3318 (2)  & -0.1833 (3)  & -0.0462 (4)  & -0.3632 (4)\\ \hline
 \multirow{4}{*}{\textbf{T5 Base}} & \textbf{Standard} & -0.0862 (2) & \textbf{0.0213 (1)}  & \textbf{-0.2372 (1)}  & -0.0853 (2)  & -0.0199 (2)  & -0.2484 (2)\\
 & \textbf{Adapter} & \textbf{-0.0844 (1)} & 0.0745 (3)  & -0.2890 (3)  & -0.1512 (3)  & \textbf{-0.0070 (1)}  & -0.2770 (3)\\
 & \textbf{Prefix} & -0.2762 (4) & -0.1256 (4)  & -0.4122 (4)  & -0.1961 (4)  & -0.0228 (3)  & -0.3983 (4)\\
 & \textbf{Last Layer} & -0.0966 (3) & 0.0280 (2)  & -0.2446 (2)  & \textbf{-0.0610 (1)}  & 0.0686 (4)  & \textbf{-0.1966 (1)}\\ \hline
 \multirow{4}{*}{\textbf{T5 Large}} & \textbf{Standard} & -0.0612 (2) & 0.0493 (3)  & \textbf{-0.2220 (1)}  & -0.0976 (2)  & \textbf{-0.0141 (1)}  & -0.2473 (2)\\
 & \textbf{Adapter} & -0.1692 (4) & -0.0042 (2)  & -0.2660 (3)  & -0.3429 (4)  & -0.1436 (4)  & -0.5082 (4)\\
 & \textbf{Prefix} & -0.1263 (3) & \textbf{0.0032 (1)}  & -0.2541 (2)  & -0.1998 (3)  & -0.0449 (2)  & -0.3452 (3)\\
 & \textbf{Last Layer} & \textbf{-0.0419 (1)} & 0.1334 (4)  & -0.2928 (4)  & \textbf{0.0050 (1)}  & 0.1224 (3)  & \textbf{-0.1843 (1)}\\ \hline
\end{tabular}%
% }
\end{adjustbox}
  \caption{Result of Second-order SPD ($\mathrm{SPD}_{2nd}$) of various models using different adaptation methods by topic, the result of model performance can be found in Appendix~\ref{sec:model_performance_by_event}. 
  }
  \label{result_table_by_topic}
  
\end{table*}

Different adaptation methods are available other than standard fine-tuning to adapt language models to a specialised task, and it has been shown that tuning a smaller set of parameters can result in more robust performance than standard fine-tuning \cite{han2021robust, li2021prefix, kirichenko2022last, chen2023unisumm}.
We investigate how different adaptation methods affect the bias introduced to the model after tuning compared to standard fine-tuning. 
% We discuss the results of both model performance and the bias introduced in this section.
We report the ROUGE 1, 2 and L scores \cite{lin-2004-rouge} and Second-order SPD mentioned in Section~\ref{the_metric} for model performance and fairness respectively.
We report Second-order SPD ($\mathrm{SPD}_{2nd}$) using different input proportions (equal, more left-leaning, and more right-leaning).
We use model performance evaluation and fairness evaluation in combination to examine adaptation methods that maintain good performance while keeping bias to a minimum level.

% The goal is to investigate if tuning less number of parameters also leads to less bias in models. Then, using model performance and the bias level, we identify the adaptation method that minimises the imposed bias while maintaining good performance.

The result can be found in Table~\ref{result_table}.
Based on ROUGE scores, we find that, not surprisingly, standard fine-tuning has the best performance since it has the highest number of parameters being updated, and adapter tuning comes second.
Depending on the model type, updating a smaller number of parameters can be less biased compared to standard fine-tune, this is especially apparent with the BART family. 
% addressing reviewer 2's 3rd question
There is a performance discrepancy between the ROUGE scores of GPT-2 models as compared to BART and T5 models. We suspect this is due to encoder-decoder language models pretrained on denoising objectives produce stronger learned representations for transfer learning \cite{patel2022bidirectional, devlin2018bert, raffel2020exploring}.
Additionally, adapter tuning has a relatively lower absolute Second-order SPD value across different input proportions compared to standard fine-tune. Combining model performance and fairness evaluation, we find that among different adaptation methods, adapter tuning has the lowest performance reduction compared to standard fine-tuning and a comparatively lower bias.

Overall, models become more left-leaning using different adaptation methods; this is witnessed by the shift of Second-order SPD for equal and more right-leaning inputs, where they have higher absolute negative values, indicating models generate summaries that expose opinions representing the left more than the right. The overall distribution of bias across various models remains similar and mainly reflects intrinsic bias.

\subsection{Different Adaptation Methods and Bias by Topic}
The dataset provided by \citet{bilal2022template} contains two topics --- COVID-19 and elections.
We divide the dataset into individual topics and fine-tune the summarisation models for each topic to investigate the effect on fairness at the single topic level.
% we further investigate different adaptation methods using various language models with each topic separately. 
All processes are the same as mentioned in Section~\ref{sec:diff_adap_methods_bias} except that we are updating models by topic separately. A detailed report and discussion of model performance can be found in Appendix~\ref{sec:model_performance_by_event}. Fairness evaluation is reported in Table~\ref{result_table_by_topic} by topic.

Similar to Section~\ref{sec:diff_adap_methods_bias}, we observe that, overall, different adaptation methods amplify bias. 
% Comparing results in Table~\ref{result_table} and Table~\ref{result_table_by_topic}, we find that standard fine-tuning on individual topics is less biased than fine-tuning using the full dataset. 
% On the other hand, b
However, by updating a smaller number of parameters, the advantage of reducing biases as opposed to using the full dataset has diminished when adapting models by topic. This suggests that when updating a smaller number of parameters, exposing the model to a narrow topic can harm the model's fairness. Indicating diversity in training data can play an important role in fairness when updating a smaller number of parameters in a model. Similar to tuning using the full dataset, models are more left-leaning using different adaptation methods by having a higher absolute negative value under equal and more right-leaning opinions provided in the input document. The overall bias distribution among models remains similar and primarily reflects intrinsic bias.

\section{Conclusion}
In this study, we examine evaluating fairness using abstractive summarisation models to summarise social media opinions, where fair models should generate summaries expose opinions from different social groups w.r.t. the provided input only. In the case of political discussion, we find that most PLMs present intrinsic bias by giving fair exposure to opinions from the left-leaning group but not the right-leaning group. We further investigate different adaptation methods and how they affect fairness.
The result shows that models adapting to the task of summarising social media text increase bias in general; however, tuning a smaller number of parameters have relatively lower bias. We further investigate tuning models by individual topic, where we find the benefit of bias reduction diminishes when tuning a smaller number of parameters, which suggests the importance of diverse datasets being presented when tuning a smaller number of parameters.
% addressing reviewer 3's last comment
Future work may explore the relationship between exposing models to diverse topics and bias.
% Out of the tested adaptation methods, adapter tuning has the lowest drop in performance while keeping bias low.
Our study sheds light on understanding bias and the effect of different adaptation methods on bias in abstractive summarisation models, particularly when summarising text with opinions. 
% We present a framework for future research on using metrics for model evaluation and fairness evaluation to evaluate model performance and bias in order to identify models that retain strong performance while minimising bias.

% In conclusion, in this study, we examined two items. First, we investigated how to measure bias using abstractive summarisation models. Second, we looked at different adaptation methods and how they affected bias in the generated summary. We found that pre-trained LLMs present intrinsic bias differently, and adaptation to the downstream task overall speaking increased bias. We further investigated the different adaptation methods' impact on bias. We found that tuning a smaller number of parameters reduced bias. Additionally, we found that out of all examined adaptation methods, adapter tuning had the lowest impact on the model's performance while having the lowest level of bias. Overall, our research highlighted the importance of understanding how bias is presented using abstractive summarisation models to summarise social media text, particularly when discussing political issues.

\section*{Limitations}
In this study, we examine bias in summarising social media text using PLMs and different adaptation methods. We focus on a single type of bias --- political bias, due to the limited dataset available.
We understand and respect the intricacies of political ideologies and recognise that they go beyond a simple binary classification. However, within the confines of our current data, categorising along the left-right spectrum provides a practical and necessary approximation for analysis. We hope that future research with more diverse datasets will allow for a more nuanced exploration of political leanings.
However, the framework of this study is applicable to different social biases in summarising social media text. 

Furthermore, due to the inability to update model parameters with different adaptation strategies in close-sourced LLMs, we focus on open-sourced language models in our work. Having stated that, the methodology for evaluating fairness using LLMs to summarise social media text is still applicable for researchers who have access to these models.

\section*{Ethics Statement}
This study followed ethical principles and guidelines.
The authors of this paper by no means suggest that language models are intentionally biased. 
% Second, it is not impossible that the authors' political leanings influenced the interpretation of the experimental result after undertaking an examination of political bias. As a result, 
We highly encourage readers to investigate and evaluate the findings for themselves. 
% Third, in the result session, we witnessed LLMs with various adaptation methods affect fairness differently. 
% We refrain from releasing these model checkpoints to avoid potential misuse of these models as a tool to sway public opinion. 
Overall, the goal of our research is to promote awareness of bias in summarising social media text since it is critical to understand what is summarised and whether it represents actual public opinion. Our work contributes to understanding the biases of summarisation models when summarising social media text, which is crucial for ethical use.

\section*{Acknowledgement}
This research is supported in part by the Australian Research Council Discovery Project DP200101441.

\bibliography{anthology,custom}

\begin{thebibliography}{60}
\expandafter\ifx\csname natexlab\endcsname\relax\def\natexlab#1{#1}\fi

\bibitem[{Amplayo et~al.(2021)Amplayo, Angelidis, and
  Lapata}]{amplayo-etal-2021-aspect}
Reinald~Kim Amplayo, Stefanos Angelidis, and Mirella Lapata. 2021.
\newblock \href {https://doi.org/10.18653/v1/2021.emnlp-main.528}
  {Aspect-controllable opinion summarization}.
\newblock In \emph{Proceedings of the 2021 Conference on Empirical Methods in
  Natural Language Processing}, pages 6578--6593, Online and Punta Cana,
  Dominican Republic. Association for Computational Linguistics.

\bibitem[{Amplayo and Lapata(2020)}]{amplayo-lapata-2020-unsupervised}
Reinald~Kim Amplayo and Mirella Lapata. 2020.
\newblock \href {https://doi.org/10.18653/v1/2020.acl-main.175} {Unsupervised
  opinion summarization with noising and denoising}.
\newblock In \emph{Proceedings of the 58th Annual Meeting of the Association
  for Computational Linguistics}, pages 1934--1945, Online. Association for
  Computational Linguistics.

\bibitem[{Ando and Zhang(2005)}]{Ando2005}
Rie~Kubota Ando and Tong Zhang. 2005.
\newblock \href {https://www.jmlr.org/papers/volume6/ando05a/ando05a.pdf} {A
  framework for learning predictive structures from multiple tasks and
  unlabeled data}.
\newblock \emph{Journal of Machine Learning Research}, 6:1817--1853.

\bibitem[{Andrew and Gao(2007)}]{andrew2007scalable}
Galen Andrew and Jianfeng Gao. 2007.
\newblock \href {https://dl.acm.org/doi/abs/10.1145/1273496.1273501} {Scalable
  training of {$L_1$}-regularized log-linear models}.
\newblock In \emph{Proceedings of the 24th International Conference on Machine
  Learning}, pages 33--40.

\bibitem[{Augenstein et~al.(2016)Augenstein, Rockt{\"a}schel, Vlachos, and
  Bontcheva}]{augenstein-etal-2016-stance}
Isabelle Augenstein, Tim Rockt{\"a}schel, Andreas Vlachos, and Kalina
  Bontcheva. 2016.
\newblock \href {https://doi.org/10.18653/v1/D16-1084} {Stance detection with
  bidirectional conditional encoding}.
\newblock In \emph{Proceedings of the 2016 Conference on Empirical Methods in
  Natural Language Processing}, pages 876--885, Austin, Texas. Association for
  Computational Linguistics.

\bibitem[{Bar-Haim et~al.(2020{\natexlab{a}})Bar-Haim, Eden, Friedman, Kantor,
  Lahav, and Slonim}]{bar2020arguments}
Roy Bar-Haim, Lilach Eden, Roni Friedman, Yoav Kantor, Dan Lahav, and Noam
  Slonim. 2020{\natexlab{a}}.
\newblock From arguments to key points: Towards automatic argument
  summarization.
\newblock In \emph{Proceedings of the 58th Annual Meeting of the Association
  for Computational Linguistics}, pages 4029--4039.

\bibitem[{Bar-Haim et~al.(2020{\natexlab{b}})Bar-Haim, Kantor, Eden, Friedman,
  Lahav, and Slonim}]{bar2020quantitative}
Roy Bar-Haim, Yoav Kantor, Lilach Eden, Roni Friedman, Dan Lahav, and Noam
  Slonim. 2020{\natexlab{b}}.
\newblock Quantitative argument summarization and beyond: Cross-domain key
  point analysis.
\newblock In \emph{Proceedings of the 2020 Conference on Empirical Methods in
  Natural Language Processing (EMNLP)}, pages 39--49.

\bibitem[{Barbieri et~al.(2020)Barbieri, Camacho-Collados, Anke, and
  Neves}]{barbieri2020tweeteval}
Francesco Barbieri, Jose Camacho-Collados, Luis~Espinosa Anke, and Leonardo
  Neves. 2020.
\newblock Tweeteval: Unified benchmark and comparative evaluation for tweet
  classification.
\newblock In \emph{Findings of the Association for Computational Linguistics:
  EMNLP 2020}, pages 1644--1650.

\bibitem[{Barocas et~al.(2019)Barocas, Hardt, and
  Narayanan}]{barocas-hardt-narayanan}
Solon Barocas, Moritz Hardt, and Arvind Narayanan. 2019.
\newblock \emph{Fairness and Machine Learning: Limitations and Opportunities}.
\newblock fairmlbook.org.
\newblock \url{http://www.fairmlbook.org}.

\bibitem[{Bilal et~al.(2022)Bilal, Wang, Tsakalidis, Nguyen, Procter, and
  Liakata}]{bilal2022template}
Iman~Munire Bilal, Bo~Wang, Adam Tsakalidis, Dong Nguyen, Rob Procter, and
  Maria Liakata. 2022.
\newblock Template-based abstractive microblog opinion summarization.
\newblock \emph{Transactions of the Association for Computational Linguistics},
  10:1229--1248.

\bibitem[{Bird et~al.(2009)Bird, Klein, and Loper}]{bird2009natural}
Steven Bird, Ewan Klein, and Edward Loper. 2009.
\newblock \emph{Natural language processing with Python: analyzing text with
  the natural language toolkit}.
\newblock " O'Reilly Media, Inc.".

\bibitem[{Blodgett et~al.(2016)Blodgett, Green, and
  O{'}Connor}]{blodgett-etal-2016-demographic}
Su~Lin Blodgett, Lisa Green, and Brendan O{'}Connor. 2016.
\newblock \href {https://doi.org/10.18653/v1/D16-1120} {Demographic dialectal
  variation in social media: A case study of {A}frican-{A}merican {E}nglish}.
\newblock In \emph{Proceedings of the 2016 Conference on Empirical Methods in
  Natural Language Processing}, pages 1119--1130, Austin, Texas. Association
  for Computational Linguistics.

\bibitem[{Bra{\v{z}}inskas et~al.(2020)Bra{\v{z}}inskas, Lapata, and
  Titov}]{bravzinskas2020unsupervised}
Arthur Bra{\v{z}}inskas, Mirella Lapata, and Ivan Titov. 2020.
\newblock Unsupervised opinion summarization as copycat-review generation.
\newblock In \emph{Proceedings of the 58th Annual Meeting of the Association
  for Computational Linguistics}, pages 5151--5169.

\bibitem[{Bra{\v{z}}inskas et~al.(2021)Bra{\v{z}}inskas, Lapata, and
  Titov}]{bravzinskas2021learning}
Arthur Bra{\v{z}}inskas, Mirella Lapata, and Ivan Titov. 2021.
\newblock Learning opinion summarizers by selecting informative reviews.
\newblock In \emph{Proceedings of the 2021 Conference on Empirical Methods in
  Natural Language Processing}, pages 9424--9442.

\bibitem[{Bra{\v{z}}inskas et~al.(2022)Bra{\v{z}}inskas, Nallapati, Bansal, and
  Dreyer}]{bravzinskas2022efficient}
Arthur Bra{\v{z}}inskas, Ramesh Nallapati, Mohit Bansal, and Markus Dreyer.
  2022.
\newblock Efficient few-shot fine-tuning for opinion summarization.
\newblock In \emph{Findings of the Association for Computational Linguistics:
  NAACL 2022}, pages 1509--1523.

\bibitem[{Center(2020)}]{pew2020differences}
Pew~Research Center. 2020.
\newblock Differences in how democrats and republicans behave on twitter.

\bibitem[{Chen and Yang(2020)}]{chen2020multi}
Jiaao Chen and Diyi Yang. 2020.
\newblock Multi-view sequence-to-sequence models with conversational structure
  for abstractive dialogue summarization.
\newblock In \emph{Proceedings of the 2020 Conference on Empirical Methods in
  Natural Language Processing (EMNLP)}, pages 4106--4118.

\bibitem[{Chen et~al.(2023)Chen, Liu, Xu, Yang, Zhu, Zeng, and
  Zhang}]{chen2023unisumm}
Yulong Chen, Yang Liu, Ruochen Xu, Ziyi Yang, Chenguang Zhu, Michael Zeng, and
  Yue Zhang. 2023.
\newblock Unisumm and summzoo: Unified model and diverse benchmark for few-shot
  summarization.
\newblock In \emph{Proceedings of the 61st Annual Meeting of the Association
  for Computational Linguistics (Volume 1: Long Papers)}, pages 12833--12855.

\bibitem[{Chu and Liu(2019)}]{chu2019meansum}
Eric Chu and Peter Liu. 2019.
\newblock Meansum: a neural model for unsupervised multi-document abstractive
  summarization.
\newblock In \emph{International Conference on Machine Learning}, pages
  1223--1232. PMLR.

\bibitem[{Dash et~al.(2019)Dash, Shandilya, Biswas, Ghosh, Ghosh, and
  Chakraborty}]{dash2019summarizing}
Abhisek Dash, Anurag Shandilya, Arindam Biswas, Kripabandhu Ghosh, Saptarshi
  Ghosh, and Abhijnan Chakraborty. 2019.
\newblock Summarizing user-generated textual content: Motivation and methods
  for fairness in algorithmic summaries.
\newblock \emph{Proceedings of the ACM on Human-Computer Interaction},
  3(CSCW):1--28.

\bibitem[{Devlin et~al.(2018)Devlin, Chang, Lee, and
  Toutanova}]{devlin2018bert}
Jacob Devlin, Ming-Wei Chang, Kenton Lee, and Kristina Toutanova. 2018.
\newblock Bert: Pre-training of deep bidirectional transformers for language
  understanding.
\newblock \emph{arXiv preprint arXiv:1810.04805}.

\bibitem[{Erkan and Radev(2004)}]{erkan2004lexrank}
G{\"u}nes Erkan and Dragomir~R Radev. 2004.
\newblock Lexrank: Graph-based lexical centrality as salience in text
  summarization.
\newblock \emph{Journal of artificial intelligence research}, 22:457--479.

\bibitem[{Fabbri et~al.(2021)Fabbri, Rahman, Rizvi, Wang, Li, Mehdad, and
  Radev}]{fabbri-etal-2021-convosumm}
Alexander Fabbri, Faiaz Rahman, Imad Rizvi, Borui Wang, Haoran Li, Yashar
  Mehdad, and Dragomir Radev. 2021.
\newblock \href {https://doi.org/10.18653/v1/2021.acl-long.535} {{C}onvo{S}umm:
  Conversation summarization benchmark and improved abstractive summarization
  with argument mining}.
\newblock In \emph{Proceedings of the 59th Annual Meeting of the Association
  for Computational Linguistics and the 11th International Joint Conference on
  Natural Language Processing (Volume 1: Long Papers)}, pages 6866--6880,
  Online. Association for Computational Linguistics.

\bibitem[{Feng et~al.(2023)Feng, Park, Liu, and
  Tsvetkov}]{feng-etal-2023-pretraining}
Shangbin Feng, Chan~Young Park, Yuhan Liu, and Yulia Tsvetkov. 2023.
\newblock \href {https://doi.org/10.18653/v1/2023.acl-long.656} {From
  pretraining data to language models to downstream tasks: Tracking the trails
  of political biases leading to unfair {NLP} models}.
\newblock In \emph{Proceedings of the 61st Annual Meeting of the Association
  for Computational Linguistics (Volume 1: Long Papers)}, pages 11737--11762,
  Toronto, Canada. Association for Computational Linguistics.

\bibitem[{Fujiwara et~al.(2021)Fujiwara, M{\"u}ller, and
  Schwarz}]{fujiwara2021effect}
Thomas Fujiwara, Karsten M{\"u}ller, and Carlo Schwarz. 2021.
\newblock The effect of social media on elections: Evidence from the united
  states.
\newblock Technical report, National Bureau of Economic Research.

\bibitem[{Goodman et~al.(2016)Goodman, Vlachos, and
  Naradowsky}]{goodman-etal-2016-noise}
James Goodman, Andreas Vlachos, and Jason Naradowsky. 2016.
\newblock \href {https://doi.org/10.18653/v1/P16-1001} {Noise reduction and
  targeted exploration in imitation learning for {A}bstract {M}eaning
  {R}epresentation parsing}.
\newblock In \emph{Proceedings of the 54th Annual Meeting of the Association
  for Computational Linguistics (Volume 1: Long Papers)}, pages 1--11, Berlin,
  Germany. Association for Computational Linguistics.

\bibitem[{Han et~al.(2021)Han, Pang, and Wu}]{han2021robust}
Wenjuan Han, Bo~Pang, and Ying~Nian Wu. 2021.
\newblock Robust transfer learning with pretrained language models through
  adapters.
\newblock In \emph{Proceedings of the 59th Annual Meeting of the Association
  for Computational Linguistics and the 11th International Joint Conference on
  Natural Language Processing (Volume 2: Short Papers)}, pages 854--861.

\bibitem[{Harper(2014)}]{harper-2014-learning}
Mary Harper. 2014.
\newblock \href {https://aclanthology.org/C14-1001} {Learning from 26
  languages: Program management and science in the babel program}.
\newblock In \emph{Proceedings of {COLING} 2014, the 25th International
  Conference on Computational Linguistics: Technical Papers}, page~1, Dublin,
  Ireland. Dublin City University and Association for Computational
  Linguistics.

\bibitem[{Harris et~al.(2020)Harris, Millman, Van Der~Walt, Gommers, Virtanen,
  Cournapeau, Wieser, Taylor, Berg, Smith et~al.}]{harris2020array}
Charles~R Harris, K~Jarrod Millman, St{\'e}fan~J Van Der~Walt, Ralf Gommers,
  Pauli Virtanen, David Cournapeau, Eric Wieser, Julian Taylor, Sebastian Berg,
  Nathaniel~J Smith, et~al. 2020.
\newblock Array programming with numpy.
\newblock \emph{Nature}, 585(7825):357--362.

\bibitem[{Hosking et~al.(2022)Hosking, Tang, and
  Lapata}]{hosking-etal-2022-hierarchical}
Tom Hosking, Hao Tang, and Mirella Lapata. 2022.
\newblock \href {https://doi.org/10.18653/v1/2022.acl-long.178} {Hierarchical
  sketch induction for paraphrase generation}.
\newblock In \emph{Proceedings of the 60th Annual Meeting of the Association
  for Computational Linguistics (Volume 1: Long Papers)}, pages 2489--2501,
  Dublin, Ireland. Association for Computational Linguistics.

\bibitem[{Hosking et~al.(2023)Hosking, Tang, and
  Lapata}]{hosking-etal-2023-attributable}
Tom Hosking, Hao Tang, and Mirella Lapata. 2023.
\newblock \href {https://doi.org/10.18653/v1/2023.acl-long.473} {Attributable
  and scalable opinion summarization}.
\newblock In \emph{Proceedings of the 61st Annual Meeting of the Association
  for Computational Linguistics (Volume 1: Long Papers)}, pages 8488--8505,
  Toronto, Canada. Association for Computational Linguistics.

\bibitem[{Houlsby et~al.(2019)Houlsby, Giurgiu, Jastrzebski, Morrone,
  De~Laroussilhe, Gesmundo, Attariyan, and Gelly}]{houlsby2019parameter}
Neil Houlsby, Andrei Giurgiu, Stanislaw Jastrzebski, Bruna Morrone, Quentin
  De~Laroussilhe, Andrea Gesmundo, Mona Attariyan, and Sylvain Gelly. 2019.
\newblock Parameter-efficient transfer learning for nlp.
\newblock In \emph{International Conference on Machine Learning}, pages
  2790--2799. PMLR.

\bibitem[{Huang et~al.(2023)Huang, Tian, Fayek, and
  Zhang}]{huang-etal-2023-examining}
Nannan Huang, Lin Tian, Haytham Fayek, and Xiuzhen Zhang. 2023.
\newblock \href {https://aclanthology.org/2023.wassa-1.14} {Examining bias in
  opinion summarisation through the perspective of opinion diversity}.
\newblock In \emph{Proceedings of the 13th Workshop on Computational Approaches
  to Subjectivity, Sentiment, {\&} Social Media Analysis}, pages 149--161,
  Toronto, Canada. Association for Computational Linguistics.

\bibitem[{Inouye and Kalita(2011)}]{inouye2011comparing}
David Inouye and Jugal~K Kalita. 2011.
\newblock Comparing twitter summarization algorithms for multiple post
  summaries.
\newblock In \emph{2011 IEEE Third international conference on privacy,
  security, risk and trust and 2011 IEEE third international conference on
  social computing}, pages 298--306. IEEE.

\bibitem[{Iso et~al.(2022)Iso, Wang, Angelidis, and
  Suhara}]{iso2022comparative}
Hayate Iso, Xiaolan Wang, Stefanos Angelidis, and Yoshihiko Suhara. 2022.
\newblock Comparative opinion summarization via collaborative decoding.
\newblock In \emph{Findings of the Association for Computational Linguistics:
  ACL 2022}, pages 3307--3324.

\bibitem[{Jakesch et~al.(2023)Jakesch, Bhat, Buschek, Zalmanson, and
  Naaman}]{jakesch2023co}
Maurice Jakesch, Advait Bhat, Daniel Buschek, Lior Zalmanson, and Mor Naaman.
  2023.
\newblock Co-writing with opinionated language models affects users’ views.
\newblock In \emph{Proceedings of the 2023 CHI Conference on Human Factors in
  Computing Systems}, pages 1--15.

\bibitem[{Johner et~al.(2021)Johner, Jana, and Biemann}]{johner2021error}
Timo Johner, Abhik Jana, and Chris Biemann. 2021.
\newblock Error analysis of using bart for multi-document summarization: A
  study for english and german language.
\newblock In \emph{Proceedings of the 23rd Nordic Conference on Computational
  Linguistics (NoDaLiDa)}, pages 391--397.

\bibitem[{Keswani and Celis(2021)}]{keswani2021dialect}
Vijay Keswani and L~Elisa Celis. 2021.
\newblock Dialect diversity in text summarization on twitter.
\newblock In \emph{Proceedings of the Web Conference 2021}, pages 3802--3814.

\bibitem[{Kirichenko et~al.(2022)Kirichenko, Izmailov, and
  Wilson}]{kirichenko2022last}
Polina Kirichenko, Pavel Izmailov, and Andrew~Gordon Wilson. 2022.
\newblock Last layer re-training is sufficient for robustness to spurious
  correlations.
\newblock In \emph{International Conference on Machine Learning}.

\bibitem[{Ladhak et~al.(2023)Ladhak, Durmus, Suzgun, Zhang, Jurafsky, Mckeown,
  and Hashimoto}]{ladhak2023pre}
Faisal Ladhak, Esin Durmus, Mirac Suzgun, Tianyi Zhang, Dan Jurafsky, Kathleen
  Mckeown, and Tatsunori~B Hashimoto. 2023.
\newblock When do pre-training biases propagate to downstream tasks? a case
  study in text summarization.
\newblock In \emph{Proceedings of the 17th Conference of the European Chapter
  of the Association for Computational Linguistics}, pages 3198--3211.

\bibitem[{Lewis et~al.(2020)Lewis, Liu, Goyal, Ghazvininejad, Mohamed, Levy,
  Stoyanov, and Zettlemoyer}]{lewis2020bart}
Mike Lewis, Yinhan Liu, Naman Goyal, Marjan Ghazvininejad, Abdelrahman Mohamed,
  Omer Levy, Veselin Stoyanov, and Luke Zettlemoyer. 2020.
\newblock Bart: Denoising sequence-to-sequence pre-training for natural
  language generation, translation, and comprehension.
\newblock In \emph{Proceedings of the 58th Annual Meeting of the Association
  for Computational Linguistics}, pages 7871--7880.

\bibitem[{Li and Liang(2021)}]{li2021prefix}
Xiang~Lisa Li and Percy Liang. 2021.
\newblock Prefix-tuning: Optimizing continuous prompts for generation.
\newblock In \emph{Proceedings of the 59th Annual Meeting of the Association
  for Computational Linguistics and the 11th International Joint Conference on
  Natural Language Processing (Volume 1: Long Papers)}, pages 4582--4597.

\bibitem[{Liang et~al.(2021)Liang, Wu, Morency, and
  Salakhutdinov}]{liang2021towards}
Paul~Pu Liang, Chiyu Wu, Louis-Philippe Morency, and Ruslan Salakhutdinov.
  2021.
\newblock Towards understanding and mitigating social biases in language
  models.
\newblock In \emph{International Conference on Machine Learning}, pages
  6565--6576. PMLR.

\bibitem[{Lin(2004)}]{lin-2004-rouge}
Chin-Yew Lin. 2004.
\newblock \href {https://aclanthology.org/W04-1013} {{ROUGE}: A package for
  automatic evaluation of summaries}.
\newblock In \emph{Text Summarization Branches Out}, pages 74--81, Barcelona,
  Spain. Association for Computational Linguistics.

\bibitem[{Liu et~al.(2019)Liu, Ott, Goyal, Du, Joshi, Chen, Levy, Lewis,
  Zettlemoyer, and Stoyanov}]{liu2019roberta}
Yinhan Liu, Myle Ott, Naman Goyal, Jingfei Du, Mandar Joshi, Danqi Chen, Omer
  Levy, Mike Lewis, Luke Zettlemoyer, and Veselin Stoyanov. 2019.
\newblock Roberta: A robustly optimized bert pretraining approach.
\newblock \emph{arXiv preprint arXiv:1907.11692}.

\bibitem[{McKinney et~al.(2011)}]{mckinney2011pandas}
Wes McKinney et~al. 2011.
\newblock pandas: a foundational python library for data analysis and
  statistics.
\newblock \emph{Python for high performance and scientific computing},
  14(9):1--9.

\bibitem[{Mihalcea and Tarau(2004)}]{mihalcea2004textrank}
Rada Mihalcea and Paul Tarau. 2004.
\newblock Textrank: Bringing order into text.
\newblock In \emph{Proceedings of the 2004 conference on empirical methods in
  natural language processing}, pages 404--411.

\bibitem[{Olabisi et~al.(2022)Olabisi, Hudson, Jetter, and
  Agrawal}]{olabisi-etal-2022-analyzing}
Olubusayo Olabisi, Aaron Hudson, Antonie Jetter, and Ameeta Agrawal. 2022.
\newblock \href {https://aclanthology.org/2022.coling-1.542} {Analyzing the
  dialect diversity in multi-document summaries}.
\newblock In \emph{Proceedings of the 29th International Conference on
  Computational Linguistics}, pages 6208--6221, Gyeongju, Republic of Korea.
  International Committee on Computational Linguistics.

\bibitem[{Paszke et~al.(2019)Paszke, Gross, Massa, Lerer, Bradbury, Chanan,
  Killeen, Lin, Gimelshein, Antiga et~al.}]{paszke2019pytorch}
Adam Paszke, Sam Gross, Francisco Massa, Adam Lerer, James Bradbury, Gregory
  Chanan, Trevor Killeen, Zeming Lin, Natalia Gimelshein, Luca Antiga, et~al.
  2019.
\newblock Pytorch: An imperative style, high-performance deep learning library.
\newblock \emph{Advances in neural information processing systems}, 32.

\bibitem[{Patel et~al.(2022)Patel, Li, Rasooli, Constant, Raffel, and
  Callison-Burch}]{patel2022bidirectional}
Ajay Patel, Bryan Li, Mohammad~Sadegh Rasooli, Noah Constant, Colin Raffel, and
  Chris Callison-Burch. 2022.
\newblock Bidirectional language models are also few-shot learners.
\newblock \emph{arXiv preprint arXiv:2209.14500}.

\bibitem[{Pedregosa et~al.(2011)Pedregosa, Varoquaux, Gramfort, Michel,
  Thirion, Grisel, Blondel, Prettenhofer, Weiss, Dubourg
  et~al.}]{pedregosa2011scikit}
Fabian Pedregosa, Ga{\"e}l Varoquaux, Alexandre Gramfort, Vincent Michel,
  Bertrand Thirion, Olivier Grisel, Mathieu Blondel, Peter Prettenhofer, Ron
  Weiss, Vincent Dubourg, et~al. 2011.
\newblock Scikit-learn: Machine learning in python.
\newblock \emph{the Journal of machine Learning research}, 12:2825--2830.

\bibitem[{Peters(2022)}]{peters2022algorithmic}
Uwe Peters. 2022.
\newblock Algorithmic political bias in artificial intelligence systems.
\newblock \emph{Philosophy \& Technology}, 35(2):25.

\bibitem[{Pfeiffer et~al.(2020)Pfeiffer, R{\"u}ckl{\'e}, Poth, Kamath, Vulic,
  Ruder, Cho, and Gurevych}]{pfeiffer2020adapterhub}
Jonas Pfeiffer, Andreas R{\"u}ckl{\'e}, Clifton Poth, Aishwarya Kamath, Ivan
  Vulic, Sebastian Ruder, Kyunghyun Cho, and Iryna Gurevych. 2020.
\newblock Adapterhub: A framework for adapting transformers.
\newblock \emph{EMNLP 2020}, page~46.

\bibitem[{Radford et~al.(2019)Radford, Wu, Child, Luan, Amodei, Sutskever
  et~al.}]{radford2019language}
Alec Radford, Jeffrey Wu, Rewon Child, David Luan, Dario Amodei, Ilya
  Sutskever, et~al. 2019.
\newblock Language models are unsupervised multitask learners.
\newblock \emph{OpenAI blog}, 1(8):9.

\bibitem[{Raffel et~al.(2020)Raffel, Shazeer, Roberts, Lee, Narang, Matena,
  Zhou, Li, Liu et~al.}]{raffel2020exploring}
Colin Raffel, Noam Shazeer, Adam Roberts, Katherine Lee, Sharan Narang, Michael
  Matena, Yanqi Zhou, Wei Li, Peter~J Liu, et~al. 2020.
\newblock Exploring the limits of transfer learning with a unified text-to-text
  transformer.
\newblock \emph{J. Mach. Learn. Res.}, 21(140):1--67.

\bibitem[{Rasooli and Tetreault(2015)}]{rasooli-tetrault-2015}
Mohammad~Sadegh Rasooli and Joel~R. Tetreault. 2015.
\newblock \href {http://arxiv.org/abs/1503.06733} {Yara parser: {A} fast and
  accurate dependency parser}.
\newblock \emph{Computing Research Repository}, arXiv:1503.06733.
\newblock Version 2.

\bibitem[{Santurkar et~al.(2023)Santurkar, Durmus, Ladhak, Lee, Liang, and
  Hashimoto}]{santurkar2023whose}
Shibani Santurkar, Esin Durmus, Faisal Ladhak, Cinoo Lee, Percy Liang, and
  Tatsunori Hashimoto. 2023.
\newblock Whose opinions do language models reflect?
\newblock \emph{arXiv preprint arXiv:2303.17548}.

\bibitem[{Sheng et~al.(2019)Sheng, Chang, Natarajan, and Peng}]{sheng2019woman}
Emily Sheng, Kai-Wei Chang, Prem Natarajan, and Nanyun Peng. 2019.
\newblock The woman worked as a babysitter: On biases in language generation.
\newblock In \emph{Proceedings of the 2019 Conference on Empirical Methods in
  Natural Language Processing and the 9th International Joint Conference on
  Natural Language Processing (EMNLP-IJCNLP)}, pages 3407--3412.

\bibitem[{Vig et~al.(2020)Vig, Gehrmann, Belinkov, Qian, Nevo, Sakenis, Huang,
  Singer, and Shieber}]{vig2020causal}
Jesse Vig, Sebastian Gehrmann, Yonatan Belinkov, Sharon Qian, Daniel Nevo,
  Simas Sakenis, Jason Huang, Yaron Singer, and Stuart Shieber. 2020.
\newblock Causal mediation analysis for interpreting neural nlp: The case of
  gender bias.
\newblock \emph{arXiv preprint arXiv:2004.12265}.

\bibitem[{Wolf et~al.(2020)Wolf, Debut, Sanh, Chaumond, Delangue, Moi, Cistac,
  Rault, Louf, Funtowicz et~al.}]{wolf2020transformers}
Thomas Wolf, Lysandre Debut, Victor Sanh, Julien Chaumond, Clement Delangue,
  Anthony Moi, Pierric Cistac, Tim Rault, R{\'e}mi Louf, Morgan Funtowicz,
  et~al. 2020.
\newblock Transformers: State-of-the-art natural language processing.
\newblock In \emph{Proceedings of the 2020 conference on empirical methods in
  natural language processing: system demonstrations}, pages 38--45.

\end{thebibliography}
\bibliographystyle{acl_natbib}

\appendix

\section{Appendix}
\subsection{Zero-shot Model Performance}
\label{sec:zero_short_model_performance}

% \begin{table}[tbh!]
%   \caption{ 
%   }
%   \label{zero_shot_performance}
% \resizebox{\columnwidth}{!}{%
% \begin{tabular}{l c c c c c c}
% \toprule
% Model & ROUGE-1 & ROUGE-2 & ROUGE-L & $\mathrm{SPD}_{2nd}$-Equal & $\mathrm{SPD}_{2nd}$-Left & $\mathrm{SPD}_{2nd}$-Right\\ \midrule
% BART Base & 22.28 & 6.49 & 15.34 & -0.0262 & 0.1219 & -0.2285\\ 
% BART Large & 22.21 & 6.56 & 14.98 & -0.0240 & 0.0708 & -0.2279\\ 
% Distil GPT-2 & 9.32 & 0.91 & 6.51 & -0.1154 & 0.0321 & -0.3520\\ 
% GPT-2 & 11.17 & 1.13 & 7.62 & -0.0345 & -0.0115 & -0.2839\\ 
% GPT-2 Medium & 10.96 & 1.26 & 7.43 & -0.0162 & -0.0160 & -0.2619\\ 
% GPT-2 Large & 10.78 & 1.29 & 7.53 & 0.0012 & -0.0345 & -0.2913\\ 
% T5 Small & 26.42 & 8.39 & 18.52 & -0.0415 & 0.0424 & -0.1957\\
% T5 Base & 26.71 & 8.48 & 19.01 & -0.1385 & -0.0390 & -0.2479\\
% T5 Large & 15.09 & 5.33 & 11.25 & -0.0160 & 0.1205 & -0.2698\\ \bottomrule
% \end{tabular}%
% }
% \end{table}

\begin{table}[tbh!]
\resizebox{\columnwidth}{!}{%
\begin{tabular}{l c c c}
\toprule
Model & ROUGE-1 & ROUGE-2 & ROUGE-L\\ \midrule
BART Base & 22.28 & 6.49 & 15.34\\ 
BART Large & 22.21 & 6.56 & 14.98\\ 
Distil GPT-2 & 9.32 & 0.91 & 6.51\\ 
GPT-2 & 11.17 & 1.13 & 7.62\\ 
GPT-2 Medium & 10.96 & 1.26 & 7.43\\ 
GPT-2 Large & 10.78 & 1.29 & 7.53\\ 
T5 Small & 26.42 & 8.39 & 18.52\\
T5 Base & 26.71 & 8.48 & 19.01\\
T5 Large & 15.09 & 5.33 & 11.25\\ \bottomrule
\end{tabular}%
}
  \caption{Model performance under zero-shot setting for summary generation using social media text.
  }
  \label{zero_shot_performance}
\end{table}

\subsection{Model Performance by Topic}
\label{sec:model_performance_by_event}

\begin{table*}[tbh!]
\centering
\resizebox{\textwidth}{!}{%
\begin{tabular}{|c|c|c|c|c|c|c|c|c|c|c|c|c|c|}
\hline
\multicolumn{2}{|c|}{  } & \multicolumn{6}{|c|}{ \textbf{In-topic} }  & \multicolumn{6}{|c|}{ \textbf{Cross-topic} } \\
\hline
\multicolumn{2}{|c|}{  } & \multicolumn{3}{|c|}{ \textbf{COVID-19} } & \multicolumn{3}{|c|}{ \textbf{Elections} } & \multicolumn{3}{|c|}{ \textbf{Train-COVID-19 Test-Elections} } & \multicolumn{3}{|c|}{ \textbf{Train-Elections Test-COVID-19}} \\
\hline
\textbf{Model} & \textbf{Adaptation Methods} & \multicolumn{1}{l|}{\textbf{ROUGE-1}} & \multicolumn{1}{l|}{\textbf{ROUGE-2}} & \multicolumn{1}{l|}{\textbf{ROUGE-L}} & \multicolumn{1}{l|}{\textbf{ROUGE-1}} & \multicolumn{1}{l|}{\textbf{ROUGE-2}} & \multicolumn{1}{l|}{\textbf{ROUGE-L}} & \multicolumn{1}{l|}{\textbf{ROUGE-1}} & \multicolumn{1}{l|}{\textbf{ROUGE-2}} & \multicolumn{1}{l|}{\textbf{ROUGE-L}} & \multicolumn{1}{l|}{\textbf{ROUGE-1}} & \multicolumn{1}{l|}{\textbf{ROUGE-2}} & \multicolumn{1}{l|}{\textbf{ROUGE-L}} \\ \hline
\multirow{4}{*}{\textbf{BART Base}} & \textbf{Standard} & 30.80 & 10.74 & 21.38 & 31.62 & 11.88 & 21.53 & 29.91 & 10.25 & 20.62 & 28.66 & 9.51 & 19.96 \\
 & \textbf{Adapter} & 30.56 & 11.14 & 22.11 & 32.32 & 11.87 & 22.50 & 29.79 & 10.60 & 21.35 & 30.40 & 10.52 & 21.82 \\ 
 & \textbf{Prefix} & 29.16 & 9.63 & 20.33 & 28.39 & 8.68 & 17.29 & 27.52 & 8.33 & 18.89 & 25.21 & 6.75 & 16.79 \\ 
 & \textbf{Last Layer} & 25.45 & 6.88 & 17.77 & 30.45 & 10.43 & 20.00 & 21.25 & 5.12 & 14.84 & 26.78 & 7.52 & 18.02 \\ \hline
\multirow{4}{*}{\textbf{BART Large}} & \textbf{Standard} & 31.67 & 11.91 & 22.36 & 31.62 & 11.05 & 21.95  & 31.07 & 10.86 & 21.42 & 29.74 & 9.63 & 20.70 \\ 
 & \textbf{Adapter} & 33.38 & 12.26 & 23.72 & 31.23 & 11.82 & 21.69 & 31.07 & 10.93 & 21.81 & 28.88 & 9.53 & 20.72 \\
 & \textbf{Prefix} & 27.61 & 9.46 & 18.24 & 27.43 & 10.00 & 17.27 & 28.69 & 8.83 & 18.30 & 24.64 & 7.57 & 16.15 \\ 
 & \textbf{Last Layer} & 28.96 & 9.46 & 19.98 & 29.43 & 9.44 & 19.92 & 29.82 & 10.09 & 19.78 & 26.14 & 7.60 & 18.16 \\ \hline
\multirow{4}{*}{\textbf{Distil GPT-2}} & \textbf{Standard} & 19.87 & 4.82 & 14.47 & 24.49 & 7.10 & 17.56 & 15.63 & 2.40 & 11.00 & 16.78 & 2.76 & 12.85 \\
 & \textbf{Adapter} & 20.98 & 5.23 & 15.69 & 23.18 & 6.23 & 17.15 & 24.52 & 6.43 & 18.4 & 23.12 & 5.85 & 17.68 \\
 & \textbf{Prefix} & 10.30 & 3.37 & 7.78 & 14.42 & 3.38 & 10.26 & 9.70 & 1.87 & 7.24 & 9.64 & 3.24 & 7.29 \\
 & \textbf{Last Layer} & 15.76 & 2.91 & 11.26 & 13.34 & 2.51 & 9.04 & 9.22 & 1.35 & 6.65 & 11.90 & 1.26 & 8.75 \\ \hline
\multirow{4}{*}{\textbf{GPT-2}} & \textbf{Standard} & 21.15 & 4.85 & 14.82 & 24.77 & 6.03 & 17.27 & 19.78 & 4.14 & 13.52 & 18.89 & 3.14 & 13.43 \\
 & \textbf{Adapter} & 21.29 & 5.57 & 15.26 & 21.52 & 5.65 & 15.05 & 24.37 & 6.13 & 17.14 & 25.34 & 9.31 & 18.07\\
 & \textbf{Prefix} & 10.57 & 3.32 & 8.01 & 17.02 & 5.67 & 13.58 & 12.72 & 3.30 & 9.98 & 10.62 & 3.71 & 8.25 \\
 & \textbf{Last Layer} & 16.17 & 2.76 & 11.08 & 20.02 & 4.71 & 15.19 & 11.11 & 1.13 & 7.83 & 13.08 & 1.77 & 9.91 \\ \hline
\multirow{4}{*}{\textbf{GPT-2 Medium}} & \textbf{Standard} & 22.01 & 4.84 & 15.11 & 24.8 & 6.94 & 17.86 & 18.78 & 3.80 & 12.99 & 18.74 & 3.76 & 13.69 \\
 & \textbf{Adapter} & 23.19 & 6.05 & 16.61 & 20.69 & 5.97 & 14.89 & 22.12 & 6.55 & 15.74 & 20.39 & 5.62 & 14.85\\
 & \textbf{Prefix} & 12.27 & 4.19 & 9.23 & 19.16 & 6.84 & 14.70 & 16.50 & 4.11 & 12.15 & 11.28 & 3.73 & 8.86 \\
 & \textbf{Last Layer} & 14.79 & 2.43 & 10.72 & 14.97 & 2.54 & 10.33 & 11.73 & 1.44 & 8.24 & 12.49 & 0.92 & 9.08 \\ \hline
\multirow{4}{*}{\textbf{GPT-2 Large}} & \textbf{Standard} & 24.77 & 7.05 & 17.98 & 25.81 & 8.20 & 18.73 & 22.12 & 6.15 & 15.64 & 21.67 & 4.74 & 15.32 \\
 & \textbf{Adapter} & 20.91 & 5.67 & 15.01 & 22.80 & 6.25 & 16.66 & 25.46 & 7.41 & 17.99 & 23.16 & 5.56 & 16.38\\
 & \textbf{Prefix} & 12.44 & 4.51 & 9.02 & 14.07 & 4.62 & 10.56 & 18.79 & 5.72 & 14.41 & 10.70 & 3.33 & 7.87 \\
 & \textbf{Last Layer} & 17.28 & 3.28 & 12.21 & 18.72 & 5.05 & 13.13 & 13.31 & 2.03 & 8.75 & 14.02 & 2.35 & 10.88 \\ \hline
\multirow{4}{*}{\textbf{T5 Small}} & \textbf{Standard} & 27.58 & 9.38 & 19.40 & 29.50 & 10.88 & 20.19 & 28.16 & 10.24 & 20.08 & 28.24 & 9.36 & 19.98\\
 & \textbf{Adapter} & 21.42 & 7.12 & 15.40 & 25.39 & 9.45 & 17.96 & 25.73 & 9.36 & 17.98 & 24.89 & 8.37 & 17.74\\ 
 & \textbf{Prefix} & 28.15 & 9.76 & 19.35 & 27.83 & 9.87 & 18.28 & 26.92 & 8.67 & 18.42 & 25.15 & 7.45 & 17.16 \\
 & \textbf{Last Layer} & 27.59 & 9.02 & 19.32 & 27.57 & 9.96 & 19.24  & 25.93 & 8.45 & 17.82 & 27.66 & 8.68 & 19.42\\ \hline
\multirow{4}{*}{\textbf{T5 Base}}& \textbf{Standard} & 28.72 & 9.55 & 20.06 & 30.19 & 10.63 & 20.44  & 29.71 & 10.34 & 20.19 & 29.76 & 9.44 & 20.67\\
 & \textbf{Adapter} & 22.94 & 8.46 & 16.35 & 28.12 & 10.20 & 19.23 & 24.93 & 8.22 & 16.90 & 25.49 & 8.32 & 17.71\\
 & \textbf{Prefix} & 24.82 & 8.18 & 18.24 & 26.98 & 9.07 & 18.12  & 25.63 & 9.01 & 17.66 & 25.67 & 8.08 & 17.53\\ 
 & \textbf{Last Layer} & 26.66 & 8.18 & 18.15 & 28.21 & 8.98 & 18.5 & 27.59 & 8.32 & 18.04 & 28.60 & 8.85 & 19.25\\ \hline
\multirow{4}{*}{\textbf{T5 Large}} & \textbf{Standard} & 30.92 & 11.26 & 21.80 & 31.29 & 12.07 & 21.88 & 30.78 & 11.63 & 22.21 & 28.86 & 9.68 & 20.01\\
 & \textbf{Adapter} & 29.61 & 10.65 & 21.06 & 30.52 & 11.57 & 21.28 & 28.88 & 10.54 & 20.99 & 28.21 & 9.16 & 19.41 \\
 & \textbf{Prefix} & 26.86 & 9.76 & 19.76 & 29.13 & 11.12 & 19.84 & 27.43 & 10.44 & 19.47 & 27.27 & 9.33 & 18.94\\
 & \textbf{Last Layer} & 18.57 & 5.90 & 13.39 & 27.81 & 9.65 & 18.81 & 21.05 & 7.79 & 15.54 & 23.75 & 7.28 & 16.60\\ \hline
\end{tabular}%
}
\caption{In the in-topic setting, for the COVID-19 partition, adapter has the overall best performance by obtaining the highest ROUGE scores; for elections, standard fine-tune has the overall best ROUGE scores. When applying models in a cross-topic setting, most models have a significant performance drop, except for those fine-tuned using adapter tuning. Suggesting adapter tuning is the most robust method for summarising social media text.}
\label{result_table_performance_by_event}
\end{table*}

The social media text summarisation dataset \cite{bilal2022template} contains two discussed topics, namely COVID-19 and elections. We divide the dataset into individual topics and train summarisation models mentioned in Section~\ref{models} using different adaptation methods mentioned in Section~\ref{adaptation_methods} for each topic separately. Then test the trained models using the provided test set by topic. In the in-topic setting, models are tested using the same topic as they are trained on, i.e., training using COVID-19 and testing using COVID-19. In the cross-topic setting, language models are tested using a different topic, i.e., training on COVID-19 and testing using elections. We measure the model performance using the ROUGE score \cite{lin-2004-rouge}, which is reported in Table~\ref{result_table_performance_by_event}.

In the in-topic setting, for the COVID-19 partition, adapter has the overall best performance by obtaining the highest ROUGE scores; for elections, standard fine-tune has the overall best ROUGE scores. When applying models in a cross-topic setting, most models have a significant performance drop, except for those fine-tuned using adapter tuning. Suggesting adapter tuning is the most robust method for summarising social media text.

\subsection{SPD and Second-order SPD}
\label{sec:spd_and_second_order_spd}
To verify the necessity to use Second-order SPD to measure bias, we conducted paired t-tests on the Observed SPD and Expected SPD across various input proportions, models, and adaptation methods.
The result is presented in Table~\ref{table:spd_t_test}. We found that a significant proportion of the differences between the Expected SPD and the Observed SPD exist. Indicating that using SPD alone is not sufficient to capture change in representation.

\begin{table*}[tbh!]
\centering
\begin{adjustbox}{max width=0.7\textwidth}
% \resizebox{\textwidth}{!}{%
\begin{tabular}{|c|c|c|c|c|}
\hline
\textbf{Model} & \textbf{Adaptation Methods} & \multicolumn{1}{c|}
{\textbf{Equal}} & \multicolumn{1}{c|}{\textbf{Left}} & \multicolumn{1}{c|}{\textbf{Right}} \\ \hline
\multirow{5}{*}{\textbf{BART Base}} & \textbf{Vanilla} &4.92*&-11.53*&-1.12\\
& \textbf{Standard} &0.04&-17.19*&-7.80*\\
 & \textbf{Adapter} & 0.65&-5.87*&-1.75\\
 & \textbf{Prefix} & -0.48&-20.91*&-11.15*\\
 & \textbf{Last Layer} & 0.90&-9.74*&-1.85\\ \hline

\multirow{5}{*}{\textbf{BART Large}} & \textbf{Vanilla}  &3.66*&-11.57*&-1.25\\
& \textbf{Standard} &-5.02*&-16.13*&-11.11*\\
 & \textbf{Adapter} & 3.20*&-8.34*&-1.79\\
 & \textbf{Prefix} & 3.60*&-16.36*&-5.55*\\
 & \textbf{Last Layer} & -3.87*&-12.96*&-5.99*\\ \hline

\multirow{5}{*}{\textbf{Distil GPT-2}} & \textbf{Vanilla}  & 3.51*&-14.8*&-4.32*\\
& \textbf{Standard} & -3.14*&-16.73*&-9.47*\\
 & \textbf{Adapter} & 1.43&-13.12*&-5.48*\\
 & \textbf{Prefix} & -3.77*&-14.8*&-11.41*\\
 & \textbf{Last Layer} & -2.32*&-13.85*&-8.11*\\ \hline

\multirow{5}{*}{\textbf{GPT-2}} & \textbf{Vanilla}  & -1.28&-7.67*&0.64\\
& \textbf{Standard} & -3.34*&-19.28*&-9.09*\\
 & \textbf{Adapter} & 1.90&-13.05*&-3.49*\\
 & \textbf{Prefix} & 3.65*&-8.2*&-1.62\\
 & \textbf{Last Layer} & -1.29&-10.85*&-4.68*\\ \hline

\multirow{5}{*}{\textbf{GPT-2 Medium}} & \textbf{Vanilla}  & -1.26&-8.84*&-1.09\\
& \textbf{Standard} & 0.84&-16.02*&-8.41*\\
 & \textbf{Adapter} & 1.52&-14.11*&-4.20*\\
 & \textbf{Prefix} & 2.35*&-10.03*&-4.00*\\
 & \textbf{Last Layer} & -0.46&-15.40*&-5.87*\\ \hline

 \multirow{5}{*}{\textbf{GPT-2 Large}} & \textbf{Vanilla}  & -1.95&-8.49*&0.39\\
& \textbf{Standard} & -0.35&-11.17*&-6.82*\\
 & \textbf{Adapter} & 0.03&-11.64*&-6.63*\\
 & \textbf{Prefix} & 1.04&-9.40*&-2.21*\\
 & \textbf{Last Layer} & -1.90&-11.98*&-4.35*\\ \hline
 
 \multirow{5}{*}{\textbf{T5 Small}} & \textbf{Vanilla} & 2.36*&-8.26*&-2.48*\\
& \textbf{Standard} & -4.03*&-14.53*&-10.71*\\
 & \textbf{Adapter} & -12.87*&-23.53*&-17.03*\\
 & \textbf{Prefix} & -5.31*&-21.99*&-14.70*\\
 & \textbf{Last Layer} & -3.06*&-23.29*&-11.36*\\ \hline
 
 \multirow{5}{*}{\textbf{T5 Base}} & \textbf{Vanilla} & -0.13&-7.16*&-5.29*\\
& \textbf{Standard} & 2.19*&-10.52*&-5.47*\\
 & \textbf{Adapter} & 5.99*&-4.39*&1.00\\
 & \textbf{Prefix} & -4.25*&-13.25*&-8.31*\\
 & \textbf{Last Layer} & 2.55*&-14.37*&-5.47*\\ \hline
 
 \multirow{5}{*}{\textbf{T5 Large}} & \textbf{Vanilla} & 5.38*&-10.63*&-1.20\\
& \textbf{Standard} & 0.91&-13.49*&-5.60*\\
 & \textbf{Adapter} & 1.02&-8.58*&-4.51*\\
 & \textbf{Prefix} & -12.74*&-20.70*&-18.67*\\
 & \textbf{Last Layer} & 3.62*&-13.82*&-3.52*\\ \hline
\end{tabular}%
% }
\end{adjustbox}
\caption{T-statistics by comparing SPD and Second-order SPD, denoted by * when p < 0.05. The results indicate that a significant proportion of the differences between the Expected SPD and the Observed SPD exist. Indicating that using SPD alone is not sufficient to capture change in representation.}
\label{table:spd_t_test}
\end{table*}

\subsection{Scientific Artifacts}
\label{sec:scientific_artifacts}
\noindent{\textbf{Open-source Packages}}
We utilise different open-source scientific artifacts in this work, including ROUGE \cite{lin-2004-rouge}, Pytorch \cite{paszke2019pytorch}, HuggingFace Transformers \cite{wolf2020transformers}, Scikit-learn \cite{pedregosa2011scikit}, NLTK \cite{bird2009natural}, Numpy \cite{harris2020array}, Pandas \cite{mckinney2011pandas}, regex.\footnote{https://docs.python.org/3/library/re.html}

\noindent{\textbf{Licenses}}
The annotation in social media opinion summarisation dataset \cite{bilal2022template} is under Attribution 4.0 International (CC BY 4.0 DEED). We have the permission to copy and redistribute the material in any medium or format for any purpose, even commercially; remix, transform, and build upon the material for any purpose, even commercially. While X (formerly known as Twitter) retains the ownership and rights of the content of the tweets.

\noindent{\textbf{Consistency with the intended use of all artifacts }}
We declare that the use of all models, datasets, or scientific artifacts in this paper aligns with their intended use.

\subsection{Computational Resources}
\label{sec:computational_resources}
All our experiments were conducted using four Nvidia A100 roughly for 90 hours in total.

\subsection{Experiment Details}
\label{sec:experiment_details}
\noindent{\textbf{Models}} In this study we use RoBERTa \cite{liu2019roberta} for classification. RoBERTa-base has 125 million parameters.
We use three language models and their variants for summarisation, namely BART \cite{lewis2020bart}, T5 \cite{raffel2020exploring}, and GPT-2 \cite{radford2019language}. 
BART-base has 140 million parameters. BART-large has 406 million parameters. T5-small has 60 million parameters. T5-base has 220 million parameters. T5-large has 770 million parameters. Distil GPT-2 has 82 million parameters.  GPT-2 has 117 million parameters. GPT-2 Medium has 345 million parameters. GPT-2 Large has 774 million parameters.

\noindent{\textbf{Hyperparameter}} For the political stance classifier, we used the Adam optimiser with a batch size of 16 and a learning rate of 1e-4 for 5 epochs with warmup steps of 2000.
 
To adapt models mentioned in Section \ref{models} to summarise social media text, we used adaptation methods mentioned in Section \ref{adaptation_methods}. For each adaptation method, we use a batch size of 16 for 10 epochs with early stopping and select the learning rate that yields the lowest validation loss.

\end{document}